\documentclass[10pt,journal,compsoc]{IEEEtran}

\usepackage{amsmath,amssymb,amsfonts}
\DeclareMathOperator*{\argmin}{arg\,min}
\usepackage{algorithmic}
\usepackage{graphicx}
\usepackage{textcomp}
\usepackage{xcolor}
\usepackage[ruled,vlined,linesnumbered]{algorithm2e}
\SetKwComment{Comment}{/* }{ */}
\usepackage{booktabs, multirow}

\usepackage[hidelinks]{hyperref}
\usepackage{amssymb}
\usepackage{pifont}
\newcommand{\cmark}{\ding{51}}
\newcommand{\xmark}{\ding{55}}
\usepackage{caption}
\usepackage{tabu} 
\usepackage{subcaption}
\usepackage{tikz} 
\ifCLASSOPTIONcompsoc
  \usepackage[nocompress]{cite}
\else
  \usepackage{cite}
\fi
\ifCLASSINFOpdf
\else
\fi
\begin{document}
\bstctlcite{IEEEexample:BSTcontrol}
%
\title{PUMA: Efficient Continual Graph Learning for Node Classification with Graph Condensation}
%
%
%
%

\author{Yilun~Liu,~\IEEEmembership{}
        Ruihong~Qiu,~\IEEEmembership{}
        Yanran~Tang,~\IEEEmembership{}
        Hongzhi~Yin,~\IEEEmembership{Senior Member,~IEEE,}
        and~Zi~Huang,~\IEEEmembership{Senior Member,~IEEE}
\IEEEcompsocitemizethanks{\IEEEcompsocthanksitem Y. Liu, R. Qiu, Y. Tang, H. Yin and Z. Huang are with the School of Electrical Engineering and Computer Science, The University of Queensland, Brisbane, Australia.\protect\\
E-mail: yilun.liu@uq.edu.au, r.qiu@uq.edu.au, yanran.tang@uq.edu.au, h.yin1@uq.edu.au, helen.huang@uq.edu.au}
\thanks{Manuscript received April 19, 2005; revised August 26, 2015.}}

%
%

\markboth{Journal of \LaTeX\ Class Files,~Vol.~14, No.~8, August~2015}%
{Shell \MakeLowercase{\textit{et al.}}: Bare Advanced Demo of IEEEtran.cls for IEEE Computer Society Journals}
%



\IEEEtitleabstractindextext{%
\begin{abstract}
When handling streaming graphs, existing graph representation learning models encounter a catastrophic forgetting problem, where previously learned knowledge of these models is easily overwritten when learning with newly incoming graphs. In response, Continual Graph Learning (CGL) emerges as a novel paradigm enabling graph representation learning from static to streaming graphs. Our prior work, Condense and Train (CaT)~\cite{cat} is a replay-based CGL framework with a balanced continual learning procedure, which designs a small yet effective memory bank for replaying data by condensing incoming graphs. Although the CaT alleviates the catastrophic forgetting problem, there exist three issues: (1) The graph condensation algorithm derived in CaT only focuses on labelled nodes while neglecting abundant information carried by unlabelled nodes; (2) The continual training scheme of the CaT overemphasises on the previously learned knowledge, limiting the model capacity to learn from newly added memories; (3) Both the condensation process and replaying process of the CaT are time-consuming. In this paper, we propose a \textbf{P}s\textbf{U}do-label guided \textbf{M}emory b\textbf{A}nk (PUMA) CGL framework, extending from the CaT to enhance its efficiency and effectiveness by overcoming the above-mentioned weaknesses and limits. To fully exploit the information in a graph, PUMA expands the coverage of nodes during graph condensation with both labelled and unlabelled nodes. Furthermore, a training-from-scratch strategy is proposed to upgrade the previous continual learning scheme for a balanced training between the historical and the new graphs. Besides, PUMA uses a one-time prorogation and wide graph encoders to accelerate the graph condensation and the graph encoding process in the training stage to improve the efficiency of the whole framework. Extensive experiments on four datasets \textcolor{black}{for the node classification task} demonstrate the state-of-the-art performance and efficiency over existing methods. The code has been released in \url{https://github.com/superallen13/PUMA}.
\end{abstract}

}

\maketitle

\IEEEdisplaynontitleabstractindextext

%
\IEEEpeerreviewmaketitle

\ifCLASSOPTIONcompsoc
\IEEEraisesectionheading{\section{Introduction}\label{sec:introduction}}
\else
\section{Introduction}
\label{sec:introduction}
\fi

%
%
%
%

\IEEEPARstart{G}{raph} representation learning aims to model the graph-structured data, typically in recommender systems~\cite{recsys-tkde}, traffic prediction~\cite{traffic-tkde} and protein function prediction~\cite{protein-tkde}. Generally, graphs are treated as static data in traditional graph representation learning, where a model is fixed once it has been trained. However, for many scenarios in the real world (e.g., social networks, citation networks and knowledge graphs), graph data are changing and evolving in a streaming manner~\cite{cgl-recsys1, cgl-kg, gag}. Recently, continual graph learning (CGL) has emerged to handle the streaming graph by adapting the static graph neural networks (GNNs). In CGL, the most significant challenge is how to address a catastrophic forgetting problem, where a model easily forgets the previously learned knowledge while overemphasising on the incoming graphs~\cite{ergnn,twp,ssm,cat}.

A few attempts have been made to tackle this catastrophic forgetting problem through leveraging regularisation penalty~\cite{twp}, architecture redesign~\cite{hpn}, and replayed graphs~\cite{ergnn, ssm}.
Among all, the replay-based methods achieve the best model performance and flexibility by storing and replaying a memory bank, which contains either informative nodes~\cite{ergnn} or subgraphs~\cite{ssm}. In our previous work, the Condensed and Train (CaT)~\cite{cat} was proposed to enhance the efficacy of memory banks and the balanced learning. The CaT consists of two main modules: a Condensed Graph Memory (CGM), which continually condenses incoming graphs to small yet informative replayed graphs, and a Train in Memory (TiM), which trains the CGL model only using the memory bank where replayed graphs have similar sizes to guarantee the balanced learning.

\begin{figure}[!t]
\centering
\includegraphics[width=\linewidth]{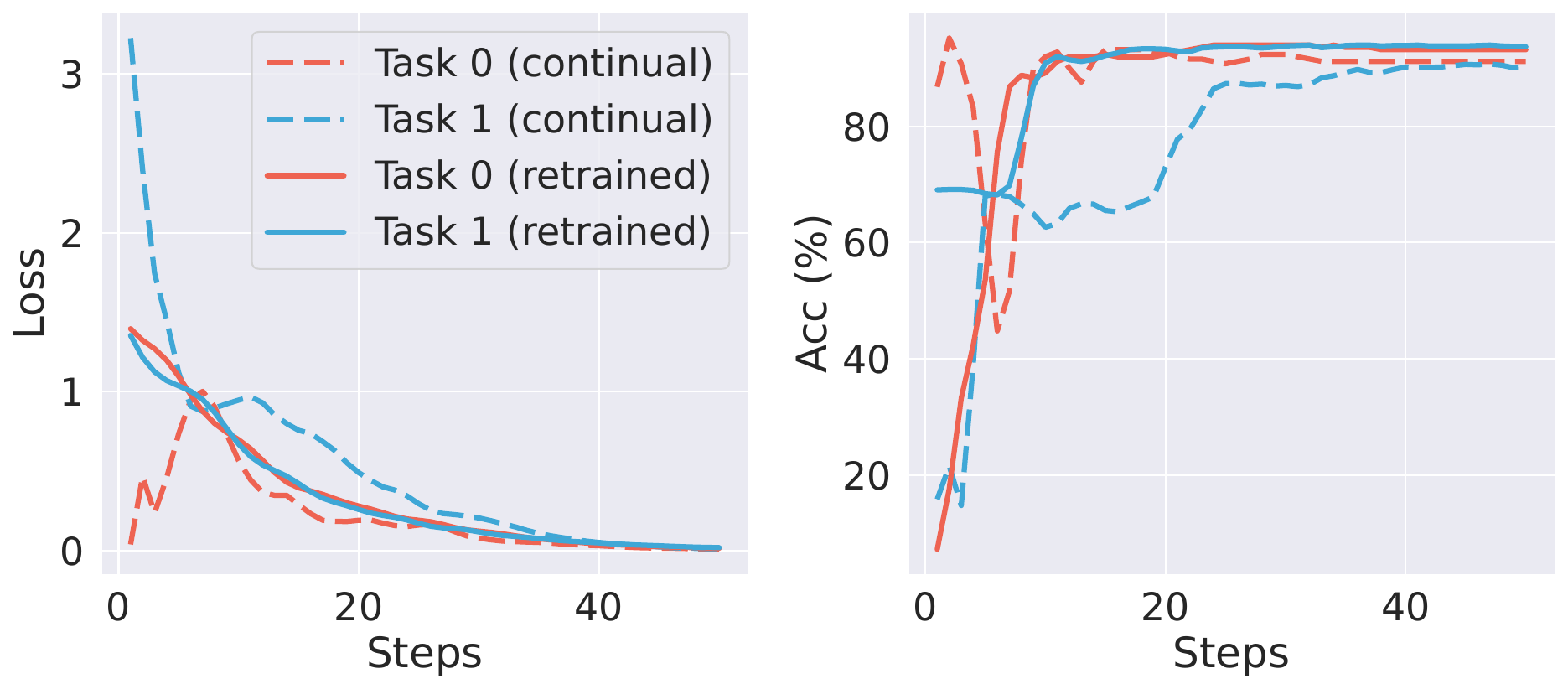}
\caption{Loss and accuracy comparisons for the first two tasks when training the replayed graph of Task 1 while replaying the replayed graph of Task 0 at the first 50 steps after the replayed graph of Task 0 has been learned 500 steps. The continual (dotted lines) and retrained (solid lines) loss values for Task 0 (red) and Task 1 (blue) are shown, providing a visual representation of an optimisation problem.}
\label{fig:vibration}
\end{figure}

Although our previous CaT framework~\cite{cat} has demonstrated that the incoming graphs can be successfully condensed to an extremely small scale, for example, 0.1\% of the original size, for the continual learning, it remains a few challenges in further improving the efficacy and efficiency. (1) \textbf{Neglected unlabelled nodes}: In the previous condensation process, nodes without label will be considered as the target for condensation rather than just supporting the message passing for the labelled nodes, potentially overlooking the valuable information from unlabelled nodes. 
(2) \textbf{Imbalanced learned knowledge}: Since previously generated memories will be repeatedly replayed during the continual learning, historical knowledge is easier to retain while the perception of new information in incoming graphs is hindered. 
Figure~\ref{fig:vibration} shows this situation for CGL that the CaT model converges faster and better for the knowledge from previous Task 0 with a smaller loss than the newly incoming Task 1. This is due to the fact that neural networks are prone to learn shortcuts from training data~\cite{shortcut}.
(3) \textbf{Low efficiency}: Both the condensation and the CGL training are inefficient, because of repetitive computations, slow convergence speed of condensation and unnecessary message passing in the GNN training. 
The condensation process requires to repeatedly map the incoming graphs to different embedding spaces, which causes a large number of recalculations for message passing among node neighbourhoods. 
Besides, condensed graph converging is impeded by narrow graph encoders (e.g., encoders with 512-dimensional hidden layers). 
Furthermore, message aggregations of neighbours on the edge-free replayed memories are redundant, which hinders the efficiency of CGL model training.

In light of the above challenges in the CaT method for CGL, this paper proposes a novel \textbf{\underline{P}se\underline{U}do-label guided \underline{M}emory b\underline{A}nk} (PUMA) CGL framework to extend the CaT in a more effective and efficient manner. Firstly, the PUMA framework develops a pseudo-labelling to integrate data from unlabelled nodes, enhancing the informativeness of the memory bank, which addresses the issue of neglected unlabelled nodes. Secondly, to tackle the inflexible historical knowledge problem, a retraining strategy is devised. This involves initialising the entire model before replaying to balance the learned knowledge among different tasks for more effective decision boundaries as well as ensuring a more stable learning process. Figure~\ref{fig:vibration} represents the optimisation process is smother, and the converge speed is quicker after using the retraining scheme. Lastly, the repetitive message passing calculation in the condensation process is simplified via a one-time propagation method, which aggregates messages from neighbouring nodes across the entire incoming graph one time, and it can be stored for reuse, significantly reducing the computational effort. Furthermore, wide graph encoders including more neurons are developed to accelerate convergence during condensation, enhancing the update efficiency of the condensed memories. Finally, to improve the training efficiency of the CGL model, the training of a multilayer perceptron (MLP) in lieu of message-passing GNNs is proposed. Due to its edge-free nature, PUMA enables the MLP to learn feature extractions and utilises a GNN to infer on the graphs with edges. These solutions collectively enhance the efficacy and efficiency of the graph condensation-based CGL framework. 

To sum up, this paper proposes a novel PUMA CGL framework extended from our previous CaT~\cite{cat} method with following substantial new contributions:
\begin{itemize}
    \item A Pseudo-label guided memory bank is proposed to condense not only labelled nodes but unlabelled nodes leveraging the information from the full graph.
    \item A retraining strategy is devised in the replay phase to alleviate the imbalanced learned knowledge problem, resulting in a promising overall performance.
    \item The condensation and the training speed is significantly improved without compromising the performance in light of the innovations consisting of a newly developed one-time propagation, wide graph encoders and MLP training with edge-free memories.
    \item Extensive experiments and in-depth analysis are conducted on four datasets, showcasing the effectiveness and the efficiency of the PUMA.
\end{itemize}

\section{Related Work}
\label{sec:liter}

\subsection{Graph Neural Networks}
Graph Neural Networks (GNNs) are effective tools for graph tasks~\cite{gin,ggnn, casegnn, fgnn}. Recent GNNs mainly employ message passing paradigm for feature representation learning~\cite{gcn,gat,sgc,sage}. \textcolor{black}{More advanced GNN architectures, such as PC-Conv~\cite{pcconv} and heterogeneous graphs~\cite{hgan,h-survey} are introduced for versatile representation learning.}

\subsection{Continual Graph Learning}
Continual graph learning is a task for handling streaming graph data in different areas~\cite{ewc,mas,gem,lwf,cgl-survey}. In graph area, CGL methods can be categorised into three branches: regularisation~\cite{twp}, replay-based~\cite{ergnn, ssm}, and architecture-based~\cite{hpn} methods. TWP~\cite{twp} preserves the topological information of historical graphs by adding regularisation. HPNs~\cite{hpn} redesigns the GNN  architecture to 3-layer prototypes for representation learning. ER-GNN~\cite{ergnn} integrates memory-replay by storing representative nodes. Sparsified subgraph memory (SSM)~\cite{ssm} stores the sparsified subgraphs by dropping edges and nodes randomly. SEM-curvature~\cite{ricci} employs balanced Forman curvature to maximise structural information within the subgraph. Structural shift risk mitigation (SSRM)~\cite{ssrm} uses regularisation-based techniques to mitigate this issue. \textcolor{black}{\cite{metaclgraph, ergnn} ensembles the graph replay and regularisation techniques to improve the model performance. Some recent work focuses on constructing advanced data structures, such as dynamic graphs~\cite{roland} and temporal graphs~\cite{dtgc, tkgr}, to solve the incremental graph problem. Compared with the existing replay-based methods~\cite{cgl-survey}, which can hardly guarantee the quality of the memory bank when the size is limited, PUMA and CaT, generate a small yet informative memory bank in a differential manner, allowing the model to learn solely from these balanced replayed graphs, thus mitigating the issue of training imbalance.} 

\subsection{Graph Condensation}
Dataset condensation generates a small and synthetic dataset to replace the original dataset and to train a model with similar performance. Dataset condensation has been applied in computer vision~\cite{dc,dcgm,dsa,cafe,dm,glad}. Recently, gradient matching has been applied to graph condensation, such as GCond~\cite{gc-multi}, DosCond~\cite{gc-one} and MCond~\cite{mcond}. DM~\cite{dm} aims to learn synthetic samples which have similar distribution with the original dataset to mimic sampling methods~\cite{herding, coreset1, coreset2}. GCDM~\cite{gc-dist} uses the distribution matching for graph condensation. SFGC~\cite{SFGC} uses a training trajectory meta-matching scheme to condense the original graph into edge-free graphs. SGDD~\cite{sgdd} broadcasts the structure information of the original graph to a condensed graph in the spectral graph theory perspective. GCondenser\cite{gcondenser} uniformly benchmarks current GC methods. Recent attempts on computer vision have directly applied dataset condensation to continual learning~\cite{cldc1, cldc2, cldc3, cldc4, cldc5}. CaT~\cite{cat} applies graph condensation to graph continual learning while ensuring the balanced training.

\section{Preliminary}
\label{sec:pre}


\subsection{Graph Neural Networks}

For a node classification problem, a graph is denoted as $\mathcal{G}=\{\boldsymbol{A}, \boldsymbol{X}, \boldsymbol{Y}\}$, where $\boldsymbol{X} \in \mathbb{R}^{n \times d}$ is the $d$-dimensional feature matrix for $n$ nodes, and the adjacency matrix $\boldsymbol{A} \in \mathbb{R}^{n \times n}$ denotes the graph structure. The graph is undirected and unweighted. $\boldsymbol{Y} \in \mathbb{R}^{n \times 1}$ are labels from a class set $\mathcal{C}$.

Graph Neural Networks (GNNs) are tools for representation learning. The node representation in GNN is calculated by message passing from neighbouring nodes as:
\begin{equation} \label{eq:gnn}
    \boldsymbol{E} = \text{GNN}_\theta(\boldsymbol{A}, \boldsymbol{X}),
\end{equation}
where $\theta$ is the model parameter and $\boldsymbol{E} \in \mathbb{R}^{n \times d^\prime}$ are $d^\prime$-dimensional node embeddings. The output $ \boldsymbol{E}^{(0)}$ of the first layer of a Graph Convolutional Network (GCN) is:
\begin{equation} \label{eq:message}
    \boldsymbol{E}^{(0)} = \sigma (\boldsymbol{L}\boldsymbol{X}\boldsymbol{W}^{(0)}),
\end{equation}
where $\boldsymbol{L} = \boldsymbol{D}^{-\frac{1}{2}}\boldsymbol{A}\boldsymbol{D}^{-\frac{1}{2}}$ is the normalised Laplacian adjacent matrix, and $\boldsymbol{D}$ is the degree matrix. $\boldsymbol{W}^{(0)} \in \mathbb{R}^{d \times d^\prime}$ is the weight of the first GCN layer. $\sigma (\cdot)$ is the activation function.

\subsection{Graph Condensation}

Graph condensation aims to synthesis a small graph $\mathcal{\tilde{G}}=\{\boldsymbol{\tilde{A}}, \boldsymbol{\tilde{X}}, \boldsymbol{\tilde{Y}}\}$ from a large graph $\mathcal{G}=\{\boldsymbol{A}, \boldsymbol{X}, \boldsymbol{Y}\}$. The model trained with the synthetic graph is expected to have a similar performance as with the original graph:
\begin{equation}
\label{eq:gc}
    \min_{\tilde{\mathcal{G}}} \mathcal{L}(\mathcal{G}; \tilde{\theta}),\quad\text{s.t.\ } \tilde{\theta} = \argmin_\theta \mathcal{L}(\tilde{\mathcal{G}}; \theta),
\end{equation}
where $\mathcal{L}$ is a task-related loss function, e.g., cross-entropy for classification problems, and $\theta$ is the parameter of GNN.

\subsection{Node Classification in Continual Graph Learning}
To solve the original node classification problem, GNN uses a linear layer to infer the node label. For example, with the embedding $\boldsymbol{E}^{(0)}$ in Eq.~\ref{eq:message}, the labels can be predicted as:
\begin{equation}
    \boldsymbol{\hat{Y}}= \text{Softmax}(\boldsymbol{L}\boldsymbol{X}\boldsymbol{E}^{(0)}\boldsymbol{W}^{(1)}).
\end{equation}

For node classification of CGL, a model is required to handle $K$ tasks $\{\mathcal{T}_1, \mathcal{T}_2,... \mathcal{T}_{K}\}$. For the $k_{th}$ task $\mathcal{T}_k$, an incoming graph $\mathcal{G}_k$ arrives and the model needs to update with $\mathcal{G}_k$ while testing on previous and the incoming tasks.

CGL problem has two different settings, task incremental learning (task-IL) and class incremental learning (class-IL). In task-IL, the model is only required to distinguish nodes in the same task. While in class-IL, the model is required to classify nodes from all tasks together. Class-IL is more challenging, and this paper focuses on this setting while also reporting the overall performance under task-IL.

\begin{figure*}[!t]
    \centering
    \includegraphics[width=0.7\linewidth]{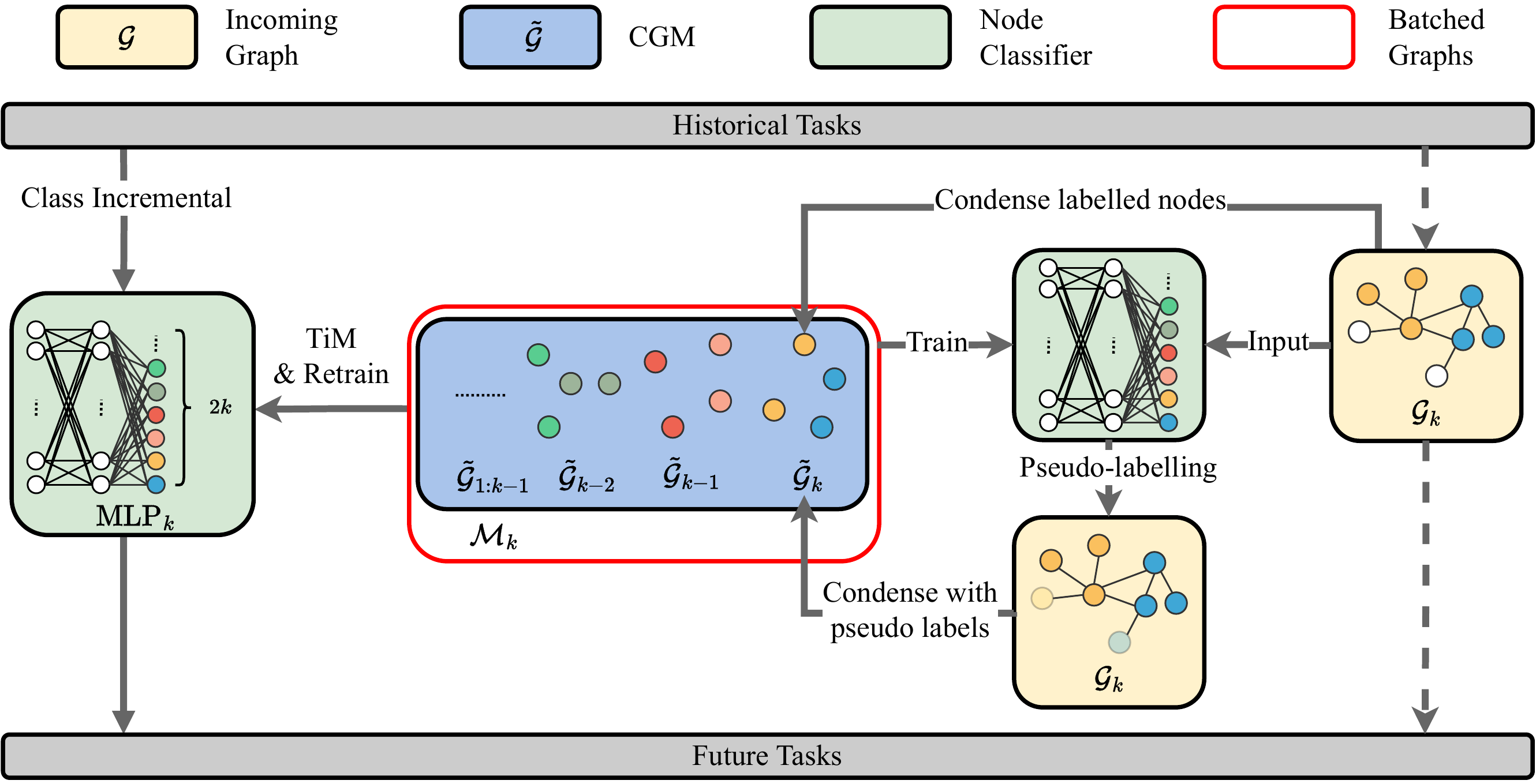}
    \caption{Details of edge-free graph condensation with pseudo-labelling and retraining. PUMA condenses the incoming graph $\mathcal{G}_{k}$ to $\mathcal{\tilde{G}}_{k}$ first, an extra classifier is trained by the memory bank $\mathcal{M}_{k}$ to assign pseudo labels to $\mathcal{G}_{k}$. After that, $\mathcal{G}_{k}$ is condensed again to update the $\mathcal{M}_{k}$. The MLP model is initialised first and trained by the $\mathcal{G}_{k}$.}
    \label{fig:puma}
\end{figure*}

\section{Methodology}
\label{sec:method}
This section describes the details of the proposed \textbf{P}se\textbf{U}do-label guided \textbf{M}emory b\textbf{A}nk (PUMA) framework. The framework details are shown in Fig.~\ref{fig:puma}.

\subsection{Fast Graph Condensation by Distribution Matching}
Condensation-based memory bank stores condensed synthetic graphs to approximate the historical data distribution. In this section, we develop an efficient graph condensation with distribution matching, aiming to maintain a similar data distribution as the original data.

For Task $\mathcal{T}_k$, the incoming graph $\mathcal{G}_k=\{\boldsymbol{A}_k,\boldsymbol{X}_k, \boldsymbol{Y}_k\}$, a edge-free condensed graph $\mathcal{\tilde{G}}_k=\{\boldsymbol{\tilde{X}}_k, \boldsymbol{\tilde{Y}}_k \}$ is generated by graph condensation. Compared with Eq.~\ref{eq:gc}, under the distribution matching scheme, the objective function of graph condensation here can be reformulated as follows:
\begin{equation}
    \mathcal{\tilde{G}}_k^* = \argmin_{\mathcal{\tilde{G}}_k} \text{Dist}(\mathcal{G}_k, \mathcal{\tilde{G}}_k),
\end{equation}
where $\text{Dist}(\cdot,\cdot)$ calculates the distance between two graphs. Using distribution matching, the distance is measured in the embedding space, where both graphs are encoded by the same encoder. CaT encodes the original graph with a specific dimensional GNN encoder and randomly initialises the encoder weights for generalisation. PUMA disentangles feature propagation and feature transformation. The incoming graph only propagates once to avoid the feature aggregation of the incoming graph repeatedly but retaining the structural information. Aggregated features $\boldsymbol{F}$ of the original graph can be calculated as pre-processing:
\begin{equation}
\boldsymbol{F}=\boldsymbol{L}\boldsymbol{X},
\end{equation}
where $\boldsymbol{L}$ is the normalised Laplacian adjacent matrix, and $\boldsymbol{X}$ is the node feature of the original graph. Feature transformation is from a linear layer with random weights $\theta$ ($\text{MLP}_{\theta}$):
\begin{align}
\label{eq:mmd}
    \mathcal{\tilde{G}}_k^* &= \argmin_{\mathcal{\tilde{G}}_k} \text{Dist}(\text{MLP}_{\theta}(\boldsymbol{\tilde{F}}), \text{MLP}_{\theta}(\boldsymbol{\tilde{X}}_k))\\
    &=\argmin_{\mathcal{\tilde{G}}_k} \text{Dist}(\boldsymbol{E}_k, \boldsymbol{\tilde{E}}_k),
\end{align}
where ${\mathcal{\tilde{G}}}_k^*=\{\boldsymbol{\tilde{X}}_k^*,\boldsymbol{\tilde{Y}}_k^*\}$ is the optimal replayed graph with distribution close to the incoming graph. Maximum mean discrepancy (MMD) is used to empirically calculate the distribution distance between two graphs. The objective is:
\begin{equation}
\label{eq:loss}
    \ell_{\text{MMD}} = \sum_{c \in \mathcal{C}_k} \frac{|\boldsymbol{E}_{c,k}|}{|\boldsymbol{E}_k|} \cdot || \text{Mean}(\boldsymbol{E}_{k,c}) - \text{Mean}(\boldsymbol{\tilde{E}}_{k,c}) ||^2,
\end{equation}
where $\mathcal{C}_k$ is the set of classes in $\mathcal{G}_k$, $\boldsymbol{E}_{c,k}$ and $\boldsymbol{\tilde{E}}_{c,k}$ are the embedding matrices of the incoming graph and condensed graph, where all nodes' labels are $c_k$. $|\cdot|$ is the number of rows in a matrix. $\text{Mean}(\cdot)$ is the mean embeddings.

To efficiently operate the condensation procedure, the feature encoders with random weights are employed here without training. The objective of the distribution matching is to minimise the embedding distance in different embedding spaces given by GNNs with random parameters $\theta_p$:
\begin{equation} \label{eq:gcdm-objective}
    \min_{\mathcal{\tilde{G}}_{k}} \sum_{\theta_p \sim \Theta} \ell_{\text{MMD},\theta_p},
\end{equation}
where $\Theta$ indicates the whole parameter space. 

With the limit of a budget as $b$, node labels $\boldsymbol{\tilde{Y}} \in \mathcal{C}_k^{b}$ for the condensed graph is initialised and kept as the same class ratio as the original graph (i.e., for any class $c_k$, $r_{k,c} \approx \tilde{r}_{k,c}$). Random sampling from the incoming graph is used to initialise the condensed node features $\boldsymbol{\tilde{X}}_k \in \mathbb{R}^{b \times d}$ at the beginning based on the assigned label. The initialisation can also be implemented as random noise.

\subsection{Pseudo Label-guided Edge-free Memory Bank}
Pseudo-labelling technique is widely used on the semi-supervised classification problem. It can enlarge the training dataset~\cite{altopt} and provide additional supervision signal~\cite{pl}. Distribution matching involves calculating class centres of training set for the original and condensed dataset in the embedding space and aligning these centres of the condensed dataset with the counterpart of original one. Therefore, unlabelled nodes with pseudo labels can be used to enlarge the training set during graph condensation. 

The unlabelled node in the incoming graph will be assigned pseudo labels by a classifier trained by the existing memory bank and the condensed memory of newly incoming graph without pseudo labels. On the other hand, adding more incorrect pseudo labels may negatively affect the distribution matching performance. Therefore, only the pseudo labels that have high confidence scores are added to the distribution matching. Logits of a node $v$ from a classifier are inputted into the Softmax function to get the confidence distribution of different classes:
\begin{equation}
    \text{confidence}(v) = \text{max}(\text{Softmax}(\textbf{A}, \textbf{X})_{[v,:]})).
\end{equation}
After obtaining confidence scores of pseudo labels, a threshold is used to filter out the uncertain pseudo labels for reducing the noisy labels. Distribution matching algorithm can enjoy the enlarged training set to accurately condense. The overall procedure of PUMA is shown in Algorithm~\ref{alg:puma}.

\begin{algorithm}[!t]
\SetAlgoLined
\caption{Pseudo-label guided memory bank (PUMA)}
\label{alg:puma}
\KwIn{Incoming graph $\mathcal{G}_k = \{\boldsymbol{A}_k,\boldsymbol{X}_k, \boldsymbol{Y}_k\}$, budget $b_k$ for the replayed graph $\mathcal{\tilde{G}}_k$, memory bank $\mathcal{M}_{k-1}$}
\KwOut{$\mathcal{\tilde{G}}_k = \{\boldsymbol{\tilde{X}}_k, \boldsymbol{\tilde{Y}}_k\}$}
Initialise labels of condensed graph $\boldsymbol{\tilde{Y}}_k$ by sampling from the label distribution of $\mathcal{G}_k$\;
Initialise node features of condensed graph $\boldsymbol{\tilde{X}}_k$ by randomly sampling $b_k$ node features from $\boldsymbol{X}_k$\;
Initialise a MLP encoder $\text{MLP}_{\theta}$ and a GNN encoder $\text{GNN}_{\theta}$ with shared parameters $\theta$\;
\For{$m \gets 1$ \KwTo $M$}{
    Initialise parameters of MLP and GNN encoders to $\theta_m$\;
    \For{$c_k \gets 0$ \KwTo $|\mathcal{C}_k|-1$}{
        $E_{c,k} = \text{GNN}_{\theta_m}(\boldsymbol{A}_k, \boldsymbol{X}_k)$\\
        $\tilde{E}_{c,k} = \text{MLP}_{\theta_m}(\boldsymbol{\tilde{X}}_k)$\\
        Calculate $\ell_{\text{MMD},\theta_m}$ according to Eq.~\ref{eq:loss}\\
        $\boldsymbol{\tilde{X}}_k \gets \boldsymbol{\tilde{X}}_k - \eta \nabla_{\boldsymbol{\tilde{X}}_k} \ell_{\text{MMD},\theta_m}$ \Comment*[r]{$\eta$ is learning rate}
    }
}

Train a classifier $\text{GNN}_{PL}$ using $\mathcal{M}_{k-1}$ and $\mathcal{\tilde{G}}_k$ to generate pseudo labels $\boldsymbol{Y}_{k,PL}$ for tested/unlabelled nodes in the $\mathcal{G}_k$\; 

\For{$n \gets 1$ \KwTo $N$}{
    Initialise parameter of MLP and GNN to $\theta_n$\;
    \For{$c_k \gets 0$ \KwTo $|\mathcal{C}_k|-1$}{
        $E_{c,k} = \text{GNN}_{\theta_n}(\boldsymbol{A}_k, \boldsymbol{X}_k)$\\
        $\tilde{E}_{c,k} = \text{MLP}_{\theta_n}(\boldsymbol{\tilde{X}}_k)$\\
        Calculate $\ell_{\text{MMD},\theta_n}$ according to Eq.~\ref{eq:loss}\\
        $\boldsymbol{\tilde{X}}_k \gets \boldsymbol{\tilde{X}}_k - \eta \nabla_{\boldsymbol{\tilde{X}}_k} \ell_{\text{MMD},\theta_n}$ \Comment*[r]{$\eta$ is learning rate}
    }
}
\end{algorithm}

ER-GNN only stores individual nodes, which can reduce the memory space and improve the replay efficiency, but ignore the structure information. SSM keeps the structure by sampling sparsified subgraphs from the incoming graphs, which improves the model's performance but does not enjoy the aggregate-free training as ER-GNN. PUMA contains edge-free graphs which can be efficiently stored in the memory and be trained by a MLP model.

\subsection{Train in Memory from Scratch}

In continual learning, the vanilla replay-based CGL methods are faced with an imbalanced learning problem. When the size of the incoming graph is significantly larger than that of replayed graphs, the model is hard to balance the learning of knowledge from the historical graphs and the incoming graph. The previous attempts for balance are based on the loss scaling, $\alpha$ and $\beta$~\cite{ergnn, ssm}, which can be represented as:
\begin{equation}
    \label{eq:rb}
    \ell_\text{replay} = \alpha\mathcal{L}(\mathcal{G}_{k};\theta_k) + \beta\mathcal{L}(\mathcal{M}_{k-1};\theta_k).
\end{equation}

In PUMA, since the condensation-based memory bank has the ability to reduce the size of a graph without compromising the performance, it is reasonable to tackle the imbalance problem by using the condensed incoming graph instead of the whole incoming graph. To incorporate this beneficial characteristic of condensed graphs into the continual learning for balanced training, when the incoming graph $\mathcal{G}_k$ arrived, the condensed graph $\mathcal{\tilde{G}}_k$ is firstly generated, which is then used to update the previous memory $\mathcal{M}_{k-1}$:
\begin{equation}
\label{eq:mem-update}
    \mathcal{M}_{k}=\mathcal{M}_{k-1} \cup \mathcal{\tilde{G}}_k.
\end{equation}

Instead of training with $\mathcal{M}_{k-1}$ and $\mathcal{G}_k$ to deal with the imbalanced issue, CaT will update the model based on $\mathcal{M}_{k}$:
\begin{equation}
\label{eq:cat}
\begin{split}
    \ell_\text{CaT} &= \mathcal{L}(\mathcal{M}_{k};\theta_k)\\
    &=\mathcal{L}(\mathcal{\tilde{G}}_{k};\theta_k) + \mathcal{L}(\mathcal{M}_{k-1};\theta_k),
\end{split}
\end{equation}
which is called Train in Memory (TiM), since the model only trains with replayed graphs in the memory bank.

On the other hand, replay-based continual learning models typically continuously update their weights instead of retraining from scratch when newly incoming graph arrives. This training scheme can encounter challenges of loss imbalance which the loss on the newly condensed graphs is larger than that of historical condensed graphs.

For an effective optimisation, the model weights of each layer are reinitialised before learning from new memories. The architecture of the CGL backbone model remains unchanged, including the number of hidden layers and hidden layer dimensions during the continual training process. 

In summary, the proposed PUMA framework uses graph condensation to generate small and effective replayed graphs, and applies the TiM scheme to solve the imbalanced learning in CGL, with details in Algorithm~\ref{alg:overall}.

\begin{algorithm}[!t]
\SetAlgoLined
\caption{Overall training procedure with PUMA}
\label{alg:overall}
\KwIn{A streaming of tasks $\{\mathcal{T}_1, \mathcal{T}_2, ..., \mathcal{T}_K\}$}
\KwOut{$\text{MLP}_{K}$}
Initialise an empty memory bank $\mathcal{M}_0$\;
\For{$k \gets 1$ \KwTo $K$}{
    Initialise a backbone model $\text{GNN}$ for CGL\;
    Initialise a MLP model shared parameters with GNN\;
    Extract incoming graph $\mathcal{G}_k$ from $\mathcal{T}_k$\;
    Obtain $\mathcal{\tilde{G}}_k$ by PUMA\Comment*[r]{Algorithm~\ref{alg:puma}} 
    $\mathcal{M}_k=\mathcal{M}_{k-1} \cup \mathcal{\tilde{G}}_k$ \Comment*[r]{Eq.~\ref{eq:mem-update}}
    Continue updating the MLP with $\mathcal{M}_k$ until the model converges \Comment*[r]{Eq.~\ref{eq:cat}}
}
\end{algorithm}

\section{Experiments}
\label{sec:exp}

\subsection{Setup}

\subsubsection{Datasets and Preprocessing}
Following the previous work~\cite{ssm,cglb}, six datasets for node classification are used for evaluation, CoraFull~\cite{corafull}, Arxiv~\cite{ogb}, Reddit~\cite{sage}, Products~\cite{ogb}, Amazon computers and Amazon photo~\cite{amazon}. Table~\ref{tab:dataset} shows the dataset statistics.

\begin{table}[!t]\centering
\caption{Dataset statistics.}\label{tab:dataset}
\resizebox{\linewidth}{!}{
\begin{tabular}{lrrrrrr}\toprule
Dataset &Nodes &Edges &Features &Classes &Tasks \\\midrule
CoraFull &19,793 &130,622 &8,710 &70 &35 \\
Arxiv &169,343 &1,166,243 &128 &40 &20 \\
Reddit &227,853 &114,615,892 &602 &40 &20 \\
Products &2,449,028 &61,859,036 &100 &46 &23 \\
\textcolor{black}{A-Computer} &13,752 &491,722 &767 &10 &5 \\ 
\textcolor{black}{A-Photo} &7,650 &238,162 &745 &8 &4\\ 
\bottomrule
\end{tabular}
}
\end{table}

Each dataset is split into a series of tasks. Each task includes nodes of two unique classes, which form an incoming graph. In each task, 60\% nodes are chosen as training nodes, 20\% nodes are for validation, and 20\% are for testing. Only the transductive setting is considered in this paper. Class-IL is the main focus of the experiment, as it is more challenging than task-IL, although overall performance in both settings is reported. In the training phase, the model can only access the newly incoming graph and the memory bank. In the testing phase, the model is required to test on all existing graphs. There are no inter-task edges between any two tasks.

\subsubsection{Baselines}
The following baselines are compared:
\begin{itemize}
    \item \textbf{Finetuning} is the lower bound baseline by updating the model only with newly incoming graphs.
    \item \textbf{Joint} is the ideal upper bound situation where the memory bank contains all incoming graphs.
    \item\textbf{EWC (2016)}~\cite{ewc} applies quadratic penalties to the model weights that are important to previous tasks.
    \item \textbf{MAS (2018)}~\cite{mas} utilises a regularisation for sensitive parameters to model performance on historical tasks.
    \item \textbf{GEM (2017)}~\cite{gem} modifies the gradients using the informative data stored in memory.
    \item \textbf{TWP (2021)}~\cite{twp} preserves the topological information for previous tasks by a regularisation term.
    \item \textbf{LwF (2018)}\cite{lwf} distils knowledge from the old model to the new model to keep the previous knowledge.
    \item \textbf{ER-GNN (2021)}~\cite{ergnn} samples the informative nodes from incoming graphs into the memory bank.
    \item \textbf{SSM (2022)}~\cite{ssm} stores the sparsified incoming graph in the memory bank for future replay.
    \item \textcolor{black}{\textbf{DSLR (2024)}~\cite{dslr} considers both the representativeness and diversity of replayed nodes.}
    \item \textbf{HPNs (2023)}~\cite{hpn} redesign the embedding generation for the task-IL by maintaining three-level prototypes.
    \item \textcolor{black}{\textbf{MetaCLGraph 
    (2023)}~\cite{metaclgraph} guides model updates by memory banks and gradients from incoming graphs.}
    \item \textcolor{black}{\textbf{ER-LS-GS (2023)}~\cite{erkd} preserves the local and the global structure of replayed graphs by regularisation.}
    \item \textbf{CaT (2023)}~\cite{cat} utilises graph condensation in CGL, which is proposed in our previous work.
\end{itemize}

\subsubsection{Evaluation Metrics} 
When the model is updated after Task $\mathcal{T}_k$, all previous tasks from $\mathcal{T}_{1}$ to $\mathcal{T}_{k}$ are evaluated. A lower triangular performance matrix $\boldsymbol{M} \in \mathbb{R}^{K \times K}$ is maintained, where $m_{i,j}$ denotes the classification accuracy of Task $\mathcal{T}_j$ after learning from Task $\mathcal{T}_i$ ($i \leq j$). Additionally, the following metrics are used to compare different methods comprehensively.

\textbf{Average performance (AP)} measures the average model performance after learning from Task $\mathcal{T}_k$:
\begin{equation}
    \text{AP}_k = \frac{1}{k}\sum^k_{i=1}m_{k,i}.
\end{equation}

\textbf{Mean of average performance (mAP)}~\cite{classIL-survey} denotes the average performance of the model in continual learning:
\begin{equation}
    \text{mAP} = \frac{1}{k}\sum^k_{i=1}\text{AP}_i.
\end{equation}

\textbf{Backward transfer (BWT)}~\cite{cl-survey1} (also known as the average forgetting (AF)) indicates how the training process of the current task affects the previous tasks. The larger number implies that training the current task will have a greater impact on historical tasks. A negative or a positive number implies a negative or a positive impact, respectively:
\begin{equation}
    \text{BWT}_k = \frac{1}{k-1}\sum_{i=1}^{k-1} (m_{k,i} - m_{i,i}).
\end{equation}

\begin{table*}[!t]\centering
\caption{Overall results for class-IL setting without inter-task edges. All replay-based CGL methods have a budget ratio of 0.005. BWT is also called average forgetting (AF). The bold results are the best performance excluding Joint, and the underlined results are the best baselines excluding Joint. $\uparrow$ denotes the greater value represents greater performance.}\label{tab:classIL}
\resizebox{\linewidth}{!}{
    \begin{tabular}{l|l|cc|cc|cc|cc|cc|cc}\toprule
    \multirow{2}{*}[-0.3ex]{Category} &\multirow{2}{*}[-0.3ex]{Methods} &\multicolumn{2}{c|}{CoraFull} &\multicolumn{2}{c|}{Arxiv} &\multicolumn{2}{c|}{Reddit} &\multicolumn{2}{c|}{Products} &\multicolumn{2}{c|}{\textcolor{black}{A-Computer}} &\multicolumn{2}{c}{\textcolor{black}{A-Photo}} \\\cmidrule{3-14}
    & &AP (\%) $\uparrow$ &BWT (\%) $\uparrow$ &AP (\%) $\uparrow$ &BWT (\%) $\uparrow$ &AP (\%) $\uparrow$ &BWT (\%) $\uparrow$ &AP (\%) $\uparrow$ &BWT (\%) $\uparrow$ &\textcolor{black}{AP (\%) $\uparrow$} &\textcolor{black}{BWT (\%) $\uparrow$} &\textcolor{black}{AP (\%) $\uparrow$} &\textcolor{black}{BWT (\%) $\uparrow$} \\\midrule
    Lower bound&Finetuning &2.2±0.0 &-96.6±0.1 &5.0±0.0 &-96.7±0.1 &5.0±0.0 &-99.6±0.0 &4.3±0.0 &-97.2±0.1 &19.6±0.2 &-99.2±0.2 &24.2±0.1 &-95.2±7.3 \\\midrule
    \multirow{4}{*}{Regularisation} &EWC &2.9±0.2 &-96.1±0.3 &5.0±0.0 &-96.8±0.1 &5.3±0.6 &-99.2±0.7 &7.6±1.1 &-91.7±1.4 &19.6±0.2 &-99.2±0.2 &24.2±0.1 &-94.5±8.1 \\
    &MAS &2.2±0.0 &-94.1±0.6 &4.9±0.0 &-95.0±0.7 &10.7±1.4 &-92.7±1.5 &10.1±0.6 &-89.0±0.5 &19.6±0.1 &-99.1±0.1 &23.5±0.8 &-98.4±0.2 \\
    &GEM &2.5±0.1 &-96.6±0.1 &5.0±0.0 &-96.8±0.1 &5.3±0.5 &-99.3±0.5 &4.3±0.1 &-96.8±0.1 &19.6±0.2 &-99.2±0.2 &24.2±0.1 &-94.5±8.1   \\
    &TWP &21.2±3.2 &-67.4±1.6 &4.3±1.1 &-93.0±8.3 &9.5±2.0 &-35.5±5.5 &6.8±3.5 &-64.3±12.8 &18.9±0.9 &-96.8±2.9 &23.0±1.2 &-98.8±0.3 \\\midrule
    Distillation &LWF &2.2±0.0 &-96.6±0.1 &5.0±0.0 &-96.8±0.1 &5.0±0.0 &-99.5±0.0 &4.3±0.0 &-96.8±0.2 &19.7±0.0 &-98.6±1.4 &23.7±1.2 &-96.9±3.7 \\\midrule
    \multirow{3}{*}{Replay} &ER-GNN &3.1±0.2 &-94.6±0.2 &23.2±0.5 &-77.1±0.5 &20.0±3.0 &-83.7±3.1 &34.0±1.0 &-55.7±0.8 &19.9±0.0 &-99.0±0.1 &24.3±0.1 &-98.7±0.1 \\
    &SSM &3.5±0.5 &-94.7±0.5 &26.4±0.8 &-73.7±0.9 &41.8±3.2 &-60.8±3.4 &58.1±0.4 &-29.3±0.5 &28.6±8.6 &-88.2±10.7 &25.7±1.9 &-96.7±2.4 \\
    &\textcolor{black}{DSLR} &3.9±0.7 &-94.2±0.4 &25.2±2.5 &-75.0±2.7 &25.0±0.4 &-51.5±0.6 &37.5±1.3 &-43.6±1.2 &24.9±0.4 &-51.7±0.7 &24.8±1.0 &-97.9±1.4 \\
    \midrule
    \textcolor{black}{\multirow{2}{*}{Ensemble}} &\textcolor{black}{MetaCLGraph} &63.8±1.5 
    &-24.7±1.6 &30.9±1.1 &-51.7±0.7 &71.9±4.4 &-28.6±4.6 &19.9±0.0 &-99.0±0.1 &29.3±19.6 &-47.1±23.5 &19.1±10.5 &-76.6±6.0 \\
    &\textcolor{black}{ER-LS-GS} &3.0±0.2 &-94.7±0.3 &24.0±0.9 &-76.3±0.9 &24.9±0.2 &-51.6±0.7 &34.1±0.8 &-55.4±0.9 &19.9±0.0 &-99.0±0.1 &24.3±0.1 &-98.6±0.1  \\
    \midrule
    Full dataset &Joint &85.3±0.1 &-2.7±0.0 &63.5±0.3 &-15.7±0.4 &98.2±0.0 &-0.5±0.0 &72.2±0.4 &-5.3±0.5 &96.4±2.1 &-1.9±2.7 &97.3±0.1 &-0.9±0.1 \\
    \midrule
    \midrule
    \multirow{2}{*}{Condensation} &CaT (ours) &\underline{68.5±0.9} &\underline{-5.7±1.3} &\underline{64.9±0.3} &\underline{-12.5±0.8} &\underline{97.7±0.1} &\underline{-0.4±0.1} &\underline{71.1±0.3} &\underline{-5.4±0.3} &\underline{90.5±1.1} &\underline{-3.1±0.7} &\underline{92.0±0.6} &\underline{-1.9±0.5} \\
    &PUMA (ours) &\textbf{77.9±0.2} &\textbf{-4.2±0.9} &\textbf{67.0±0.1} &\textbf{-12.2±0.3} &\textbf{98.0±0.1} &\textbf{-0.3±0.1} &\textbf{74.2±0.4} &\textbf{-4.1±0.5} &\textbf{91.7±0.4} &\textbf{-2.6±0.2} &\textbf{93.5±0.5} &\textbf{-1.8±0.5} \\
    \bottomrule
    \end{tabular}
}
\end{table*}

\begin{table*}[!t]\centering
\caption{Overall results in task-IL setting without inter-task edges. $^*$The results of HPNs are from the original paper.}\label{tab:taskIL}
\resizebox{\linewidth}{!}{
    \begin{tabular}{l|l|cc|cc|cc|cc|cc|cc}\toprule
    \multirow{2}{*}[-0.3ex]{Category} &\multirow{2}{*}[-0.3ex]{Methods} &\multicolumn{2}{c|}{CoraFull} &\multicolumn{2}{c|}{Arxiv} &\multicolumn{2}{c|}{Reddit} &\multicolumn{2}{c|}{Products} &\multicolumn{2}{c|}{\textcolor{black}{A-Computer}} &\multicolumn{2}{c}{\textcolor{black}{A-Photo}} \\
    \cmidrule{3-14}
    & &AP (\%) $\uparrow$ &BWT (\%) $\uparrow$ &AP (\%) $\uparrow$ &BWT (\%) $\uparrow$ &AP (\%) $\uparrow$ &BWT (\%) $\uparrow$ &AP (\%) $\uparrow$ &BWT (\%) $\uparrow$ &\textcolor{black}{AP (\%) $\uparrow$} &\textcolor{black}{BWT (\%) $\uparrow$} &\textcolor{black}{AP (\%) $\uparrow$} &\textcolor{black}{BWT (\%) $\uparrow$} \\\midrule
    Lower bound&Finetuning &51.0±3.4 &-46.2±3.5 &67.1±5.2 &-31.3±5.6 &57.1±7.4 &-44.6±7.8 &56.4±3.8 &-42.4±4.0 &48.7±12.4 &-62.5±14.9 &71.6±10.1 &-32.0±17.2 \\\midrule
    \multirow{4}{*}{Regularisation} &EWC &87.4±2.2 &-9.1±2.2 &85.6±7.7 &-11.9±8.1 &85.5±3.3 &-14.8±3.5 &90.3±1.8 &-6.8±1.9 &75.8±5.3 &-29.0±6.6 &74.7±7.1 &-27.2±10.5 \\
    &MAS &93.0±0.3 &-0.7±0.5 &83.8±6.9 &-12.0±7.8 &99.0±0.1 &\textbf{0.0±0.0} &\textbf{95.9±0.1} &0.0±0.0 &77.7±11.0 &-26.6±13.6 &82.6±11.5 &-19.7±14.8 \\
    &GEM &\underline{94.3±0.6} &-2.1±0.5 &\underline{94.7±0.1} &-2.3±0.2 &99.3±0.1 &-0.3±0.1 &86.9±0.9 &-10.6±0.9 &78.96±8.4 &-22.59±9.6 &93.05±3.5 &-6.35±2.3 \\
    &TWP &87.9±1.9 &-4.9±0.6 &77.1±7.3 &-3.5±5.4 &74.1±5.5 &-1.5±0.5 &75.5±4.4 &-4.9±6.4 &79.3±13.5 &-23.1±16.2 &86.0±7.9 &-13.3±11.8 \\\midrule
    Distillation &LwF &64.7±1.1 &-32.3±1.2 &60.2±5.8 &-38.6±6.2 &62.4±3.5 &-39.1±3.7 &50.1±0.7 &-49.3±0.8 &78.2±9.1 &-26.2±11.3 &89.1±5.0 &-11.4±7.0 \\\midrule
    Architecture &$\text{HPNs}^*$ &- &- &85.8±0.7 &\textbf{0.6±0.9} &- &- &80.1±0.8 &\textbf{2.9±1.0} &- &-&- &- \\
    \midrule
    \multirow{3}{*}{Replay} &ER-GNN &47.9±1.9 &-49.1±2.0 &87.3±0.8 &-10.2±0.8 &87.5±2.5 &-12.8±2.7 &80.5±1.3 &-17.3±1.3 &71.0±11.3 &-35.2±14.1 &73.6±14.0 &-32.9±18.6 \\
    &SSM &73.4±2.4 &-22.8±2.5 &91.8±0.4 &-5.5±0.5 &94.0±2.7 &-5.9±2.8 &92.3±0.8 &-4.8±0.8 &92.4±3.0 &-8.5±3.7 &95.2±1.3 &-4.1±1.6 \\
    &\textcolor{black}{DSLR} &71.7±3.3 &-24.5±3.2 &91.8±0.4 &-5.6±0.4 &49.8±1.5 &-1.9±1.9 &48.3±0.8 &-4.9±1.3 &79.4±11.5 &-24.7±14.4 &92.3±5.0 &-8.0±6.4 \\
    \midrule
    \textcolor{black}{\multirow{2}{*}{Ensemble}} &\textcolor{black}{MetaCLGraph} &92.5±0.9
    &-10.0±0.9 &84.8±1.1 &-5.5±1.1 &91.3±1.7 &-3.2±1.1 &92.5±2.3 &-4.1±0.9 &80.0±16.3 &-11.9±12.1 &69.1±16.7 &-18.2±22.6 \\
    &\textcolor{black}{ER-LS-GS} &47.4±2.3 &-49.5±2.4 &87.7±0.8 &-9.9±0.8 &52.2±1.3 &-3.0±3.3 &80.3±1.5 &-17.4±1.4 &68.4±9.9 &-38.4±12.4 &62.4±7.0 &-47.9±9.3 \\
    \midrule
    Full dataset &Joint &97.2±0.0 &0.2±0.1 &96.7±0.0 &-0.1±0.1 &99.7±0.0 &0.0±0.0 &95.7±0.7 &-0.2±0.7 &98.9±0.1 &-0.0±0.1 &98.3±0.0 &0.1±0.1 \\
    \midrule
    \midrule
    \multirow{2}{*}{Condensation} &CaT (ours) &93.3±0.4 &\textbf{-0.3±0.6} &94.7±0.3 &-0.8±0.3 &\underline{99.3±0.0} &\underline{-0.0±0.1} &94.9±0.3 &-0.5±0.5 &97.7±0.2 &-0.2±0.2 &\underline{96.5±1.5} &\textbf{0.4±0.4}  \\
    &PUMA (ours) &\textbf{95.2±0.3} &\underline{-0.7±0.2} &\textbf{95.3±0.1} &\underline{0.1±0.1} &\textbf{99.4±0.0} &\textbf{0.0±0.0} &\underline{95.4±0.3} &\underline{0.1±0.5} &\textbf{97.8±0.1} &\textbf{-0.1±0.1} &\textbf{97.3±0.2} &\underline{-0.2±0.2} \\
    \bottomrule
    \end{tabular}
    }
\end{table*}

\subsubsection{Implementation} The budget ratio represents the proportion of the memory bank to the total number of nodes in the entire training set, and the budget for every task is evenly assigned. By default, the budget ratio for the Joint baseline is 1 as it stores every training data in its memory. Unless otherwise specified, for the replay-based method, the default budget ratio is 0.005, which means the size of the memory bank becomes 0.5\% of the size of the entire training data. Although the budget is set to a real number instead of a ratio of the entire training set in more piratical scenarios, the budget ratio is used in the experiments for keeping fairness and comparing the efficiency of different memory banks.

In the condensation phase, a 1-layer MLP with 4096 dimensions is used to encode all datasets by default. The learning rate for updating condensed features is 0.001. In the continual training phase, a 3-layer MLP with two 512-dimensional hidden layers and a class number-dependent output layer is used as the CGL model for condensation-based methods. In contrast, other replay-based methods utilise a 3-layer GCN with 512 hidden dimensions. All results are obtained by running five times and reported with the average value and the standard error, and all experiments are conducted on one NVIDIA A100 GPU (80GB).

\subsection{Overall Results}
\label{sec:overall-results}
The CaT and PUMA are compared with all baselines in both class-IL and task-IL settings. AP is used to evaluate the average model performance of all learned tasks at the end of the task streaming, and BWT denotes the forgetting problem during continual learning. Table~\ref{tab:classIL} shows the overall performance of all methods in the class-IL CGL setting. CaT and PUMA achieve state-of-the-art performance compared with all other CGL baselines and can match the ideal Joint performance in the Arxiv, Reddit and Products by only maintaining a synthetic memory bank with a 0.005 budget ratio. Besides, the results show that a condensation-based memory bank has a smaller BWT, which means that the condensation not only preserves the historical knowledge but reduces the negative effects on the previous tasks while training the current task to alleviate the catastrophic forgetting problem. PUMA and CaT outperform other baselines in CoraFull but do not reach the Joint performance with two potential reasons: (1) the 0.005 budget ratio for CoraFull limits the replayed graph to only two nodes, which is extremely small to contain sufficient information; (2) CoraFull has 35 tasks, which is more than other datasets and difficult to retain historical knowledge. \textcolor{black}{The same issue of limited node sizes occurs in both the A-Computer dataset (8 nodes per replayed graph) and the A-Photo dataset (6 nodes per replayed graph), which leads to a little performance gap when compared with the joint method.}

Other baselines can hardly match the performance of condensation-based memory bank. Finetuning is easy to forget the previous knowledge since it only uses the newly incoming graph to update the model. Regularisation-based methods (e.g., EWC, MAS, GEM and TWP) also have unsatisfactory performance since adding overhead restrictions to the model will lead to decreased model plasticity during the long streaming of tasks. As a distillation method, LwF hardly handles the class-IL setting in the CGL. ER-GNN does not have reasonable results in all benchmarks for the sampling-based replay methods since there is a severe imbalanced training problem. SSM stores sparsified subgraphs in the memory bank, which can preserve the topological information for the historical graph data. Although SSM has a good performance, it still has a gap to Joint or condensation-based methods. \textcolor{black}{DSLR enhances the diversity of the memory bank but does not significantly improve model performance due to budget limitations and imbalance training issues. Ensemble methods that combine the memory bank with the regularisation techniques (e.g., MetaCLGraph and ER-LS-GS) still suffer inadequate historical information and imbalanced training problems.}

Table~\ref{tab:taskIL} shows the overall performance under the task-IL setting. Compared to class-IL, task-IL is much easier, and all baselines achieve a reasonable result. The CaT and PUMA can match the Joint method with a 0.005 budget ratio.

\subsection{Ablation Study}
The PUMA has two key components, pseudo-label guided memory bank and retraining. To study the effectiveness, different variants are evaluated, and mAP of these variants is reported in Table~\ref{tab:ablation} to indicate overall performance. The variant without PL indicates that only labelled nodes are used in the condensation process, while the variant without Re implies that the CGL model updates its weights based on the learned knowledge from previous tasks. According to Table~\ref{tab:ablation}, compared to the variant lacking both components, the variant using PL improves the performance in the CoraFull, Arxiv, and Products datasets, but shows no change in the Reddit dataset. This is because the labelled data in CoraFull, Arxiv, and Products do not sufficiently capture the node feature distribution, whereas the labelled nodes in the Reddit dataset are sufficiently informative. Additionally, the variant with retraining enhances overall performance, particularly in smaller datasets such as CoraFull and Arxiv.

\begin{table}[!t]
\centering
\caption{Ablation study of the PUMA framework. mAP of different variants on all datasets.}\label{tab:ablation}
\begin{tabular}{c|c|c|c|c|c}\toprule
PL &Re &CoraFull &Arxiv &Reddit &Products \\\midrule
\xmark &\xmark &79.4±0.8 &74.7±0.3 &98.7±0.0 &83.8±0.2 \\
\xmark &\cmark &83.7±0.5 &75.6±0.1 &98.9±0.0 &84.2±0.1 \\
\cmark &\xmark &80.6±0.5 &75.7±0.3 &98.7±0.0 &84.3±0.2 \\
\cmark &\cmark &84.4±0.5 &76.4±0.3 &98.9±0.0 &84.6±0.1 \\
\bottomrule
\end{tabular}
\end{table}

\subsection{Efficacy of Condensation-based Memory Banks}
To analyse the Efficacy of condensation-based memory banks (e.g., CaT and PUMA), four memory banks are evaluated with four budget ratios, i.e., 0.005, 0.01, 0.05, and 0.1. For a fair comparison with PUMA, the TiM and retraining are applied for ER-GNN, SSM and CaT.

\begin{figure}[h]
\centering
\includegraphics[width=0.8\linewidth]{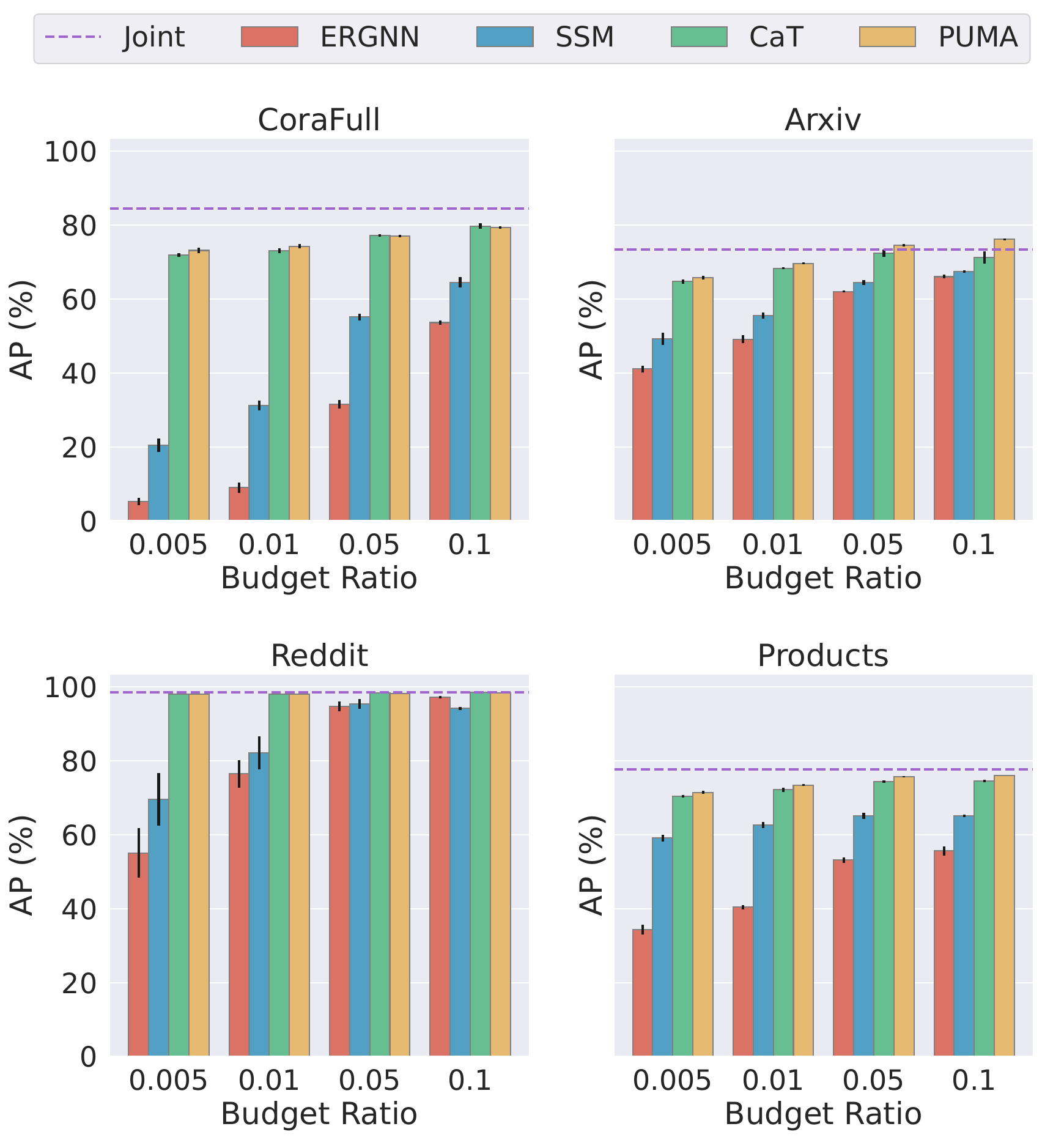}
\caption{AP of different memory banks with different budget ratios. All memory banks use both TiM and retraining to obtain the best performance for fairness.}
\label{fig:budgets}
\end{figure}


\subsubsection{Budget Efficiency}
The advantage of the condensed graph is to keep the information of the original graph while reducing the graph size. Fig.~\ref{fig:budgets} shows that the CaT and PUMA outperform the sampling-based methods with a more limited budget. In all scenarios, the sampling-based methods have a huge performance gap to CaT and PUMA. PUMA with 0.005-budget ratio can outperform or match the 0.1-budget ratio sampling-based methods on CoraFull, Reddit and Products.

On one hand, CaT and PUMA use less memory space to approximate the historical data. On the other, in the training phase, a smaller memory bank can benefit the model with storage and computation efficiency during message passing.

\subsubsection{\textcolor{black}{Label Distribution of Memory Bank}}
\textcolor{black}{Although TiM ensures that the training graphs have a balanced size, the number of nodes in each class remains proportional to the incoming graphs in the default setting. Consequently, if the incoming graph exhibits imbalanced class sizes, this imbalance will be reflected in the training graphs. To investigate how label distribution impacts condensed graph memory, we studied the effects of condensing the incoming graph with a balanced label distribution. Table~\ref{tab:label-dist} presents the performance of CaT and PUMA on the Arxiv dataset using different label distribution strategies, specifically proportional and balanced distributions. The results demonstrate that a balanced label distribution results in a performance decline. This decline occurs because the condensation ratio increases for the larger classes while it relaxes for the smaller classes compared to the proportional label distribution. This finding suggests that larger classes, which contain more information, benefit from being allocated a larger number of nodes.} 
\begin{table}[!t]
    \caption{\textcolor{black}{CaT and PUMA with different label initialisation strategies on Arxiv. Proportional label distribution means each replayed graph has 30 nodes, with the size of each class proportional to the incoming graph. In contrast, balanced label distribution ensures each class has 15 nodes.}}
    \centering
    \begin{tabular}{c|c|c|c}\toprule
        Method &Label Distribution &AP &BWT \\\midrule
        \multirow{2}{*}{CaT}& Proportional &65.4±1.9 &-12.4±1.5 \\
        & Balanced &63.6±1.3 &-13.0±1.1 \\\midrule
        \multirow{2}{*}{PUMA} & Proportional &68.4±0.2 &-10.9±0.4 \\
         & Balanced &66.5±0.5 &-10.8±0.7 \\
        \bottomrule
    \end{tabular}
    \label{tab:label-dist}
\end{table}

\subsection{Balanced Learning with TiM}
\subsubsection{Replay-based CGL Methods with TiM} TiM is a plug-and-play training scheme for replay-based CGL methods. Table ~\ref{tab:tim} shows the mAP for different replay-based CGL methods with and without TiM. It demonstrates that the TiM scheme can improve the average performance with all memory bank generation methods. The reason is that the TiM can ensure training graphs for the CGL models have a similar size to deal with the imbalanced issue, which can mitigate the catastrophic forgetting problem. 

\begin{table}[!t]
\centering
\caption{Comparison of mAP (\%) for replay-based CGL methods without and with TiM.}
\label{tab:tim}
\begin{tabular}{c|c|c|c|c|c}
\toprule
Method & TiM & CoraFull & Arxiv & Reddit & Products \\
\midrule
\multirow{2}{*}{ER-GNN} & \xmark & 11.3±0.0 & 37.6±0.6 & 50.9±1.9 & 37.1±0.6 \\
& \cmark & 13.8±0.7 & 56.5±1.1 & 77.5±1.0 & 42.0±0.5 \\
\midrule
\multirow{2}{*}{SSM} & \xmark & 11.7±0.2 & 40.6±1.1 & 70.2±0.9 & 62.1±0.3 \\
& \cmark & 27.5±3.5 & 62.8±1.4 & 77.3±3.3 & 75.9±0.4 \\
\midrule
\multirow{2}{*}{CaT} & \xmark & 29.2±1.4 & 44.1±1.2 & 89.4±0.7 & 72.7±0.2 \\
& \cmark & 76.7±0.6 & 74.0±0.4 & 98.7±0.0 & 83.7±0.2 \\
\midrule
\multirow{2}{*}{PUMA} & \xmark & 70.6±0.5 & 57.7±0.4 & 97.6±0.0 & 72.1±0.1 \\
& \cmark & 80.6±0.5 & 75.7±0.3 & 98.7±0.0 & 84.3±0.2 \\
\bottomrule
\end{tabular}
\end{table}

\subsubsection{Visualisation} 
\begin{figure}[!t]
\centering
\includegraphics[width=\linewidth]{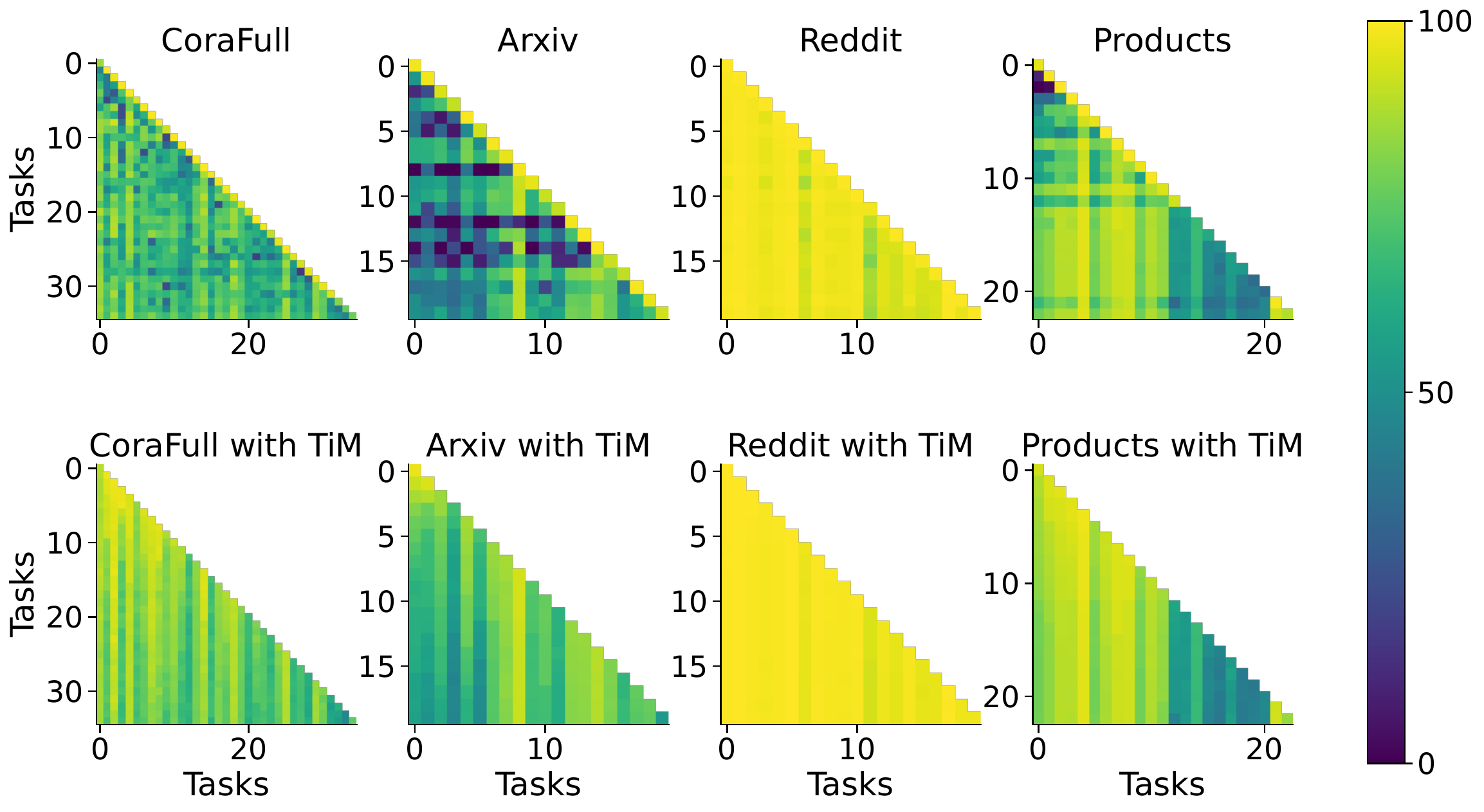}
\caption{Performance matrix visualisation of PUMA with and without TiM scheme in all four datasets. The coloured square located at the $i_{th}$ row and the $j_{th}$ column denotes the classification accuracy of Task $\mathcal{T}_j$ after model training on Task $\mathcal{T}_i$. Light colour means high accuracy, and dark colour means low accuracy. The $i_{th}$ column from top to bottom can represent the accuracy changes during the model's continual training of Task $\mathcal{T}_i$.}
\label{fig:acc}
\end{figure}

The performance matrices of PUMA on the CoraFull, Arxiv, Reddit and Products datasets under 0.01 budget ratio are visualised in Fig.~\ref{fig:acc}. All memory banks without TiM struggle with remembering the previous knowledge since the scale gap between the newly incoming graph and replayed graphs in the memory bank. After using the TiM scheme, the performance matrices show the forgetting process slows down (i.e., the colour of each column is not changed a lot), which indicates the catastrophic forgetting problem is alleviated as the imbalanced training issue is tackled.

\subsection{Effectiveness of Retraining}
Although TiM effectively mitigates the issue of training graph imbalance, it introduces a new challenge during continual training: imbalance in task losses. Memories added earlier to the memory bank are sufficiently learned, resulting in smaller losses compared to those added later. To alleviate this problem, a retraining strategy is proposed to train the model in the memories from scratch.


\begin{table}[!t]
\centering
\caption{Comparison of mAP (\%) for replay-based CGL methods without and with retraining.}
\label{tab:retrain}
\begin{tabular}{c|c|c|c|c|c}
\toprule
Method & Re & CoraFull & Arxiv & Reddit & Products \\
\midrule
\multirow{2}{*}{ER-GNN} & \xmark & 13.8±0.7 & 56.5±1.1 & 77.5±1.0 & 42.0±0.5 \\
& \cmark & 16.1±0.5 & 58.7±0.8 & 81.1±3.1 & 43.9±0.7 \\
\midrule
\multirow{2}{*}{SSM} & \xmark & 27.5±3.5 & 62.8±1.4 & 77.3±3.3 & 75.9±0.4 \\
& \cmark & 38.6±4.7 & 64.6±1.3 & 83.6±4.3 & 76.4±0.3 \\
\midrule
\multirow{2}{*}{CaT} & \xmark & 76.7±0.6 & 74.0±0.4 & 98.7±0.0 & 83.7±0.2 \\
& \cmark & 83.6±0.5 & 75.4±0.3 & 98.9±0.0 & 84.2±0.2 \\
\midrule
\multirow{2}{*}{PUMA} & \xmark & 80.6±0.5 & 75.7±0.3 & 98.7±0.0 & 84.3±0.2 \\
& \cmark & 84.4±0.5 & 76.4±0.3 & 98.9±0.0 & 84.6±0.1 \\
\bottomrule
\end{tabular}
\end{table}

\subsubsection{Effectiveness} Table~\ref{tab:retrain} represents the mAP of different replay-based methods without and with retraining. The metric mAP is selected as it enables a comprehensive evaluation of accuracy at each step of the continual learning process, rather than focusing solely on the final step performance. 

In the experiment, the efficacy of retraining for replay-based CGL methods is clearly demonstrated across various datasets. During continual training, models are prone to overfitting on previous memories. Additionally, the issue of imbalanced loss exacerbates this tendency, hindering the model's ability to effectively learn the decision boundaries between newly added and historical memories. Consequently, retraining becomes crucial, as it recalibrates the model's knowledge and adaptability to both new and existing data, thereby improving overall performance.


\subsubsection{Visualisation}

\begin{figure}[!t]
    \centering
    \includegraphics[width=0.9\linewidth]{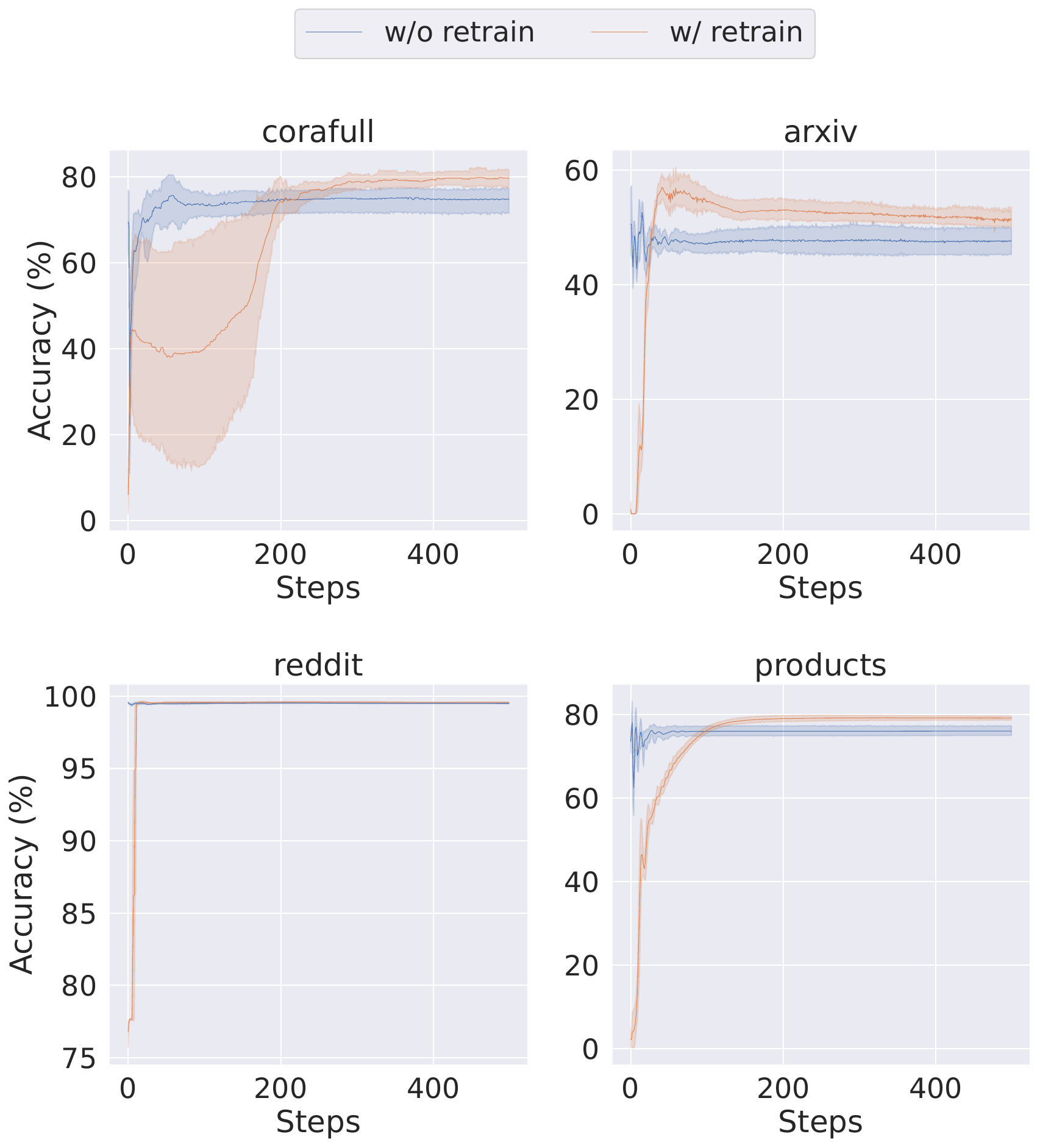}
    \caption{The accuracy changes of the first task when training with the last task on four datasets.}
    \label{fig:retrain}
\end{figure}

 Fig.~\ref{fig:retrain} shows the accuracy changes of the first task after receiving the last task of the graph streaming during continual training. For the CoraFull, Arxiv and Products datasets, the accuracy without retraining has a shake and converge then. Although the retrained model needs more steps to converge, the accuracy with retraining can outperform the continual training. For the Reddit dataset, retraining the CGL model cannot benefit a lot since the model accuracy is already high.

\subsection{Condense More by Pseudo-Labelling}
The impact of pseudo-labelling on the accuracy of replay-based CGL methods, such as ER-GNN, SSM, CaT and PUMA, is evaluated in this study. The comparison involves scenarios with and without the integration of pseudo labels during the memorisation stage. Table~\ref{tab:pl} presents the mAP for different CGL methods under both conditions.

It is observed that for condensation-based graph memory, incorporating pseudo-labelling proves to be effective. This approach aids in training a robust classifier, which in turn generates useful pseudo-labels. Additionally, introducing pseudo labels into incoming graphs expands the training dataset, providing a more comprehensive set of supervisory signals. Such an expansion significantly enhances the generalisation capabilities of the CGL model, leading to improved accuracy in inferring labels of test nodes. However, it is noted that the pseudo-labelling technique does not yield similar benefits for current sampling-based memory banks. The limitation primarily arises due to the reliance of pseudo-labelling classifiers on memory banks for training.

\begin{table}[!t]\centering
\caption{mAP of replay-based CGL methods with and without pseudo-labelling.}\label{tab:pl}
\begin{tabular}{c|c|c|c|c|c}\toprule
Method &PL &CoraFull &Arxiv &Reddit &Products \\\midrule
\multirow{2}{*}{ER-GNN} &\xmark &11.3±0.4 &51.6±1.0 &67.6±3.7 &36.2±1.7 \\
&\cmark &11.2±0.5 &50.1±0.4 &68.7±2.7 &34.5±0.8 \\\midrule
\multirow{2}{*}{SSM} &\xmark &28.0±3.5 &58.2±1.7 &75.2±6.9 &72.5±0.5 \\
&\cmark &25.4±4.9 &55.3±1.8 &75.0±4.1 &71.1±0.4 \\\midrule
\multirow{2}{*}{PUMA} &\xmark &82.8±0.7 &74.1±0.1 &98.9±0.0 &84.1±0.0 \\
&\cmark &84.1±0.6 &74.7±0.3 &98.9±0.0 &84.6±0.0 \\
\bottomrule
\end{tabular}
\end{table}

\subsection{Wide Graph Encoder}
\begin{figure}[!t]
\begin{tabular}{cc}
800-dim embeddings & 12800-dim embeddings \\
  \includegraphics[width=40mm]{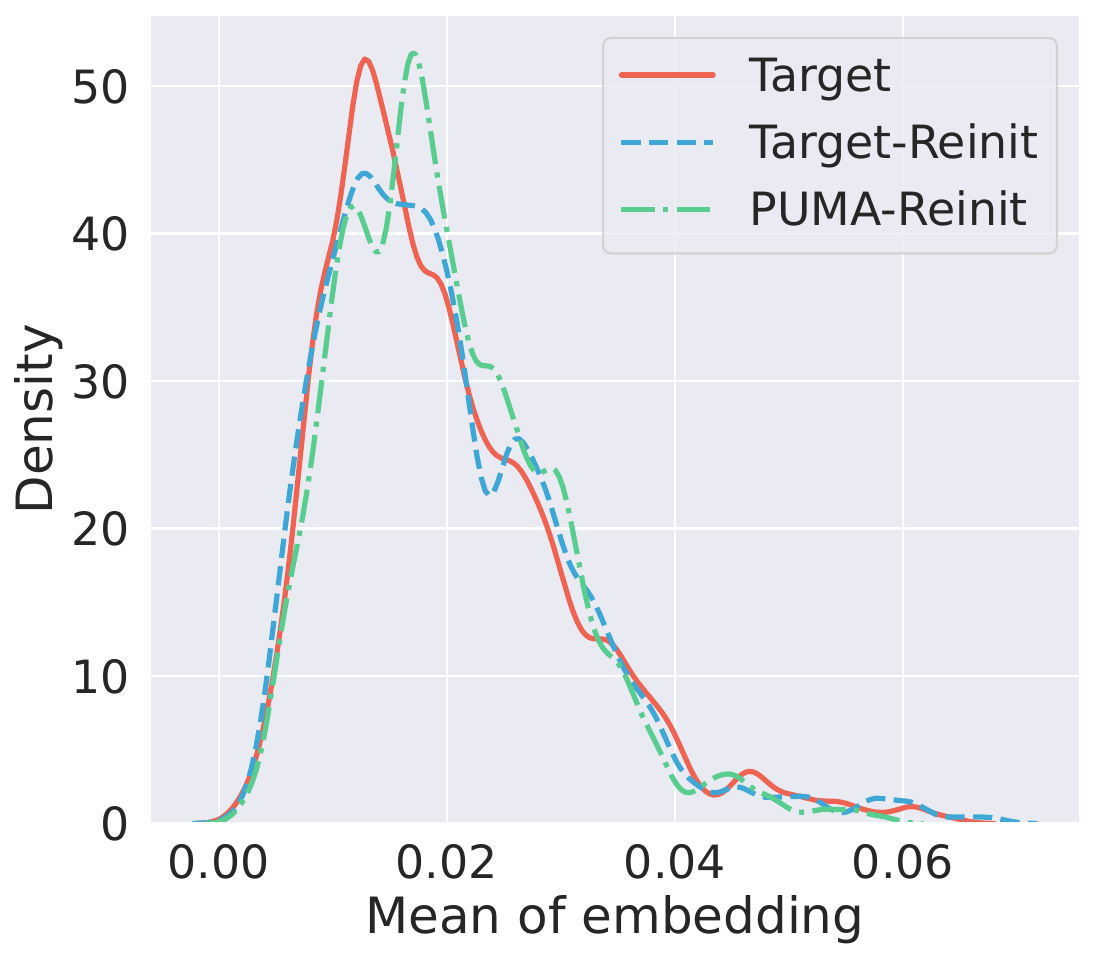} &   \includegraphics[width=40mm]{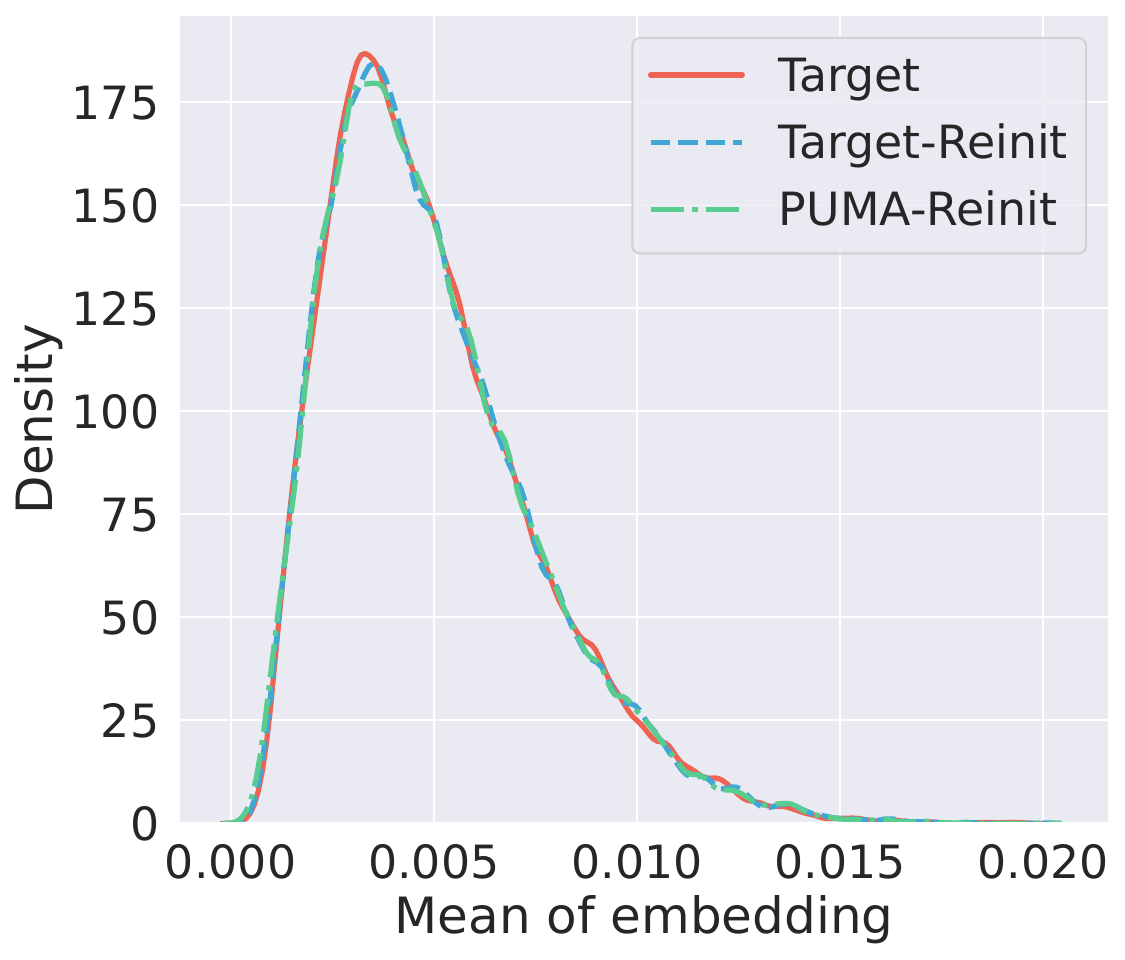}
\end{tabular}
\caption{Kernel density estimation (KDE) plot of mean embeddings. It compares KDE of mean embeddings at different stages: `Target' (original graph's embedding), `Target-Reinit' (original graph's embedding by a reinitialised encoder), and `PUMA-Reinit' (condensed graph's embedding by a reinitialised encoder). The encoder dimensions of the left figure and right figure are 800 and 12800, respectively.}
\label{fig:digits}
\end{figure}

A wide graph encoder with randomly initialised weights can randomly extract features with non-linearity. As shown in Fig.~\ref{fig:digits}, reducing the distance between the original and condensed graph embeddings generated by a narrow encoder is not sufficient, as there remains a clear distribution gap between them when the encoder is reinitialised and this gap is reduced by using wider graph encoders.

The more neurons are involved at one time, the clearer the potential transformations of the original data in the initialisation space will be, making it easier to fit the data distribution through different networks. Having an accurate data distribution is crucial in continual learning because new classes continuously emerge, and during the replay process, the model needs to relearn the decision boundary between different classes. However, using more neurons requires more computational resources, so multiple random encoders are used in practice.

\subsection{Parameter Sensitivity}
There are several hyperparameters in condensed memory bank generation, including the budget ratio for the replayed graph, which is already evaluated in Fig.~\ref{fig:budgets}. This section will discuss the choices of the dimension of graph encoders and the usage of activation during encoding.

\subsubsection{Different Dimensional Graph Encoders} 
\begin{figure}[!t]
    \centering
    \includegraphics[width=0.9\linewidth]{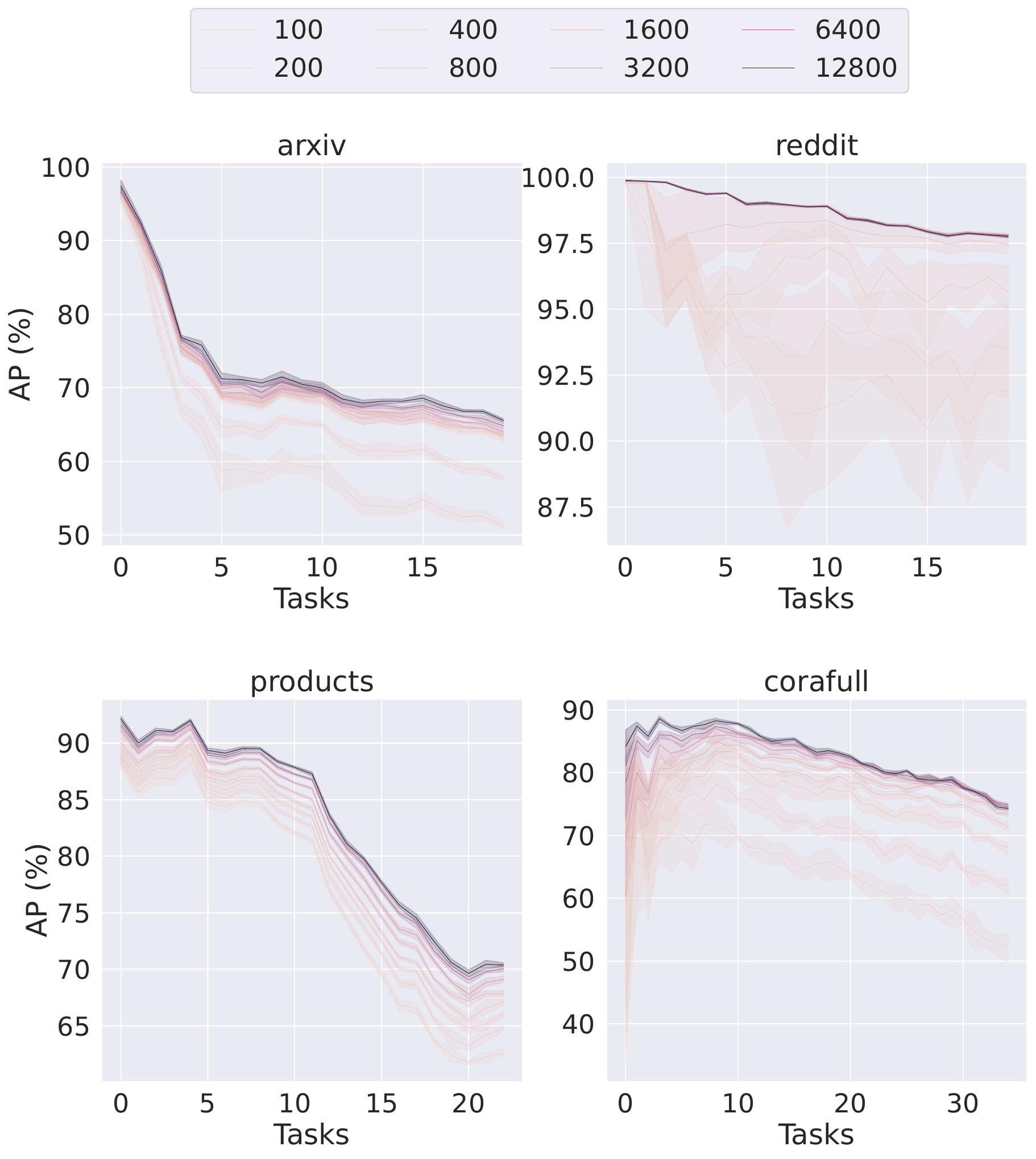}
    \caption{The average performance (AP) changes in the whole continual learning process with memory banks generated by different dimensional graph encoders. Except for the CoraFull dataset, all other datasets only update the condensed graph 500 times under only one random parameterised graph encoder. For CoraFull, 50 random graph encoders are used to update the condensed graph 100 times individually for reducing the distribution distance in a larger embedding space since CoraFull's 8710-dimensional features imply a larger model parameter initialisation space.}
    \label{fig:dim}
\end{figure}

This experiment explores the sufficient dimensions of graph encoders to optimise the MMD loss. The condense iterations remain 500, and the graph encoder is only randomly initialised once before condensation. Figure~\ref{fig:dim} shows the AP of the model trained by the PUMA generated by various dimensional graph encoders after handling each incoming task during the continual learning process. 

When the encoder dimensions are restricted because of the limited computational resources, model performance decreases since the model's parameter initialisation space is not complete. As the dimension of the graph encoder increases, the CGL model's performance experiences a significant improvement. It shows that for better covering the initialisation space of model parameters, wide encoders are essential. When the computation resources are limit, use more random relative small graph encoders can also get matching performance but need more condensation time. 

\subsubsection{Neuron Activation}
This experiment investigates the impact of using activation functions in encoders for CGL. The chosen activation function is ReLU. mAP is employed to measure effectiveness. Table~\ref{tab:encoder} illustrates that for varying datasets, encoders with activation functions are more competitive options for encoding graphs overall. It can be concluded that using ReLU can benefit the performance of CGL.


\begin{table}[!t]
\centering
\caption{mAP (\%) of different graph encoders with varying budget ratios}\label{tab:encoder}
\begin{subtable}{\linewidth}
\centering
\begin{tabular}{c|cccc}\toprule
\multirow{2}{*}{Activation} & \multicolumn{4}{c}{CoraFull} \\
\cmidrule{2-5}
& 0.005 & 0.01 & 0.05 & 0.1 \\
\midrule
\xmark & 86.4±0.0 & 58.4±0.3 & 84.4±0.4 & 86.6±0.2 \\
\cmark & 84.1±0.6 & 74.5±0.4 & 85.7±0.1 & 86.6±0.1 \\
\end{tabular}
\end{subtable}

\begin{subtable}{\linewidth}
\centering
\begin{tabular}{c|cccc}\toprule
\multirow{2}{*}{Activation} & \multicolumn{4}{c}{Arxiv} \\
\cmidrule{2-5}
& 0.005 & 0.01 & 0.05 & 0.1 \\
\midrule
\xmark & 71.4±0.5 & 73.8±0.2 & 76.7±0.3 & 77.6±0.2 \\
\cmark & 74.7±0.3 & 76.5±0.2 & 79.8±0.1 & 81.0±0.1 \\
\end{tabular}
\end{subtable}

\begin{subtable}{\linewidth}
\centering
\begin{tabular}{c|cccc}\toprule
\multirow{2}{*}{Activation} & \multicolumn{4}{c}{Reddit} \\
\cmidrule{2-5}
& 0.005 & 0.01 & 0.05 & 0.1 \\
\midrule
\xmark & 98.8±0.0 & 98.8±0.0 & 98.9±0.0 & 98.9±0.0 \\
\cmark & 98.9±0.0 & 99.0±0.0 & 99.2±0.0 & 99.2±0.0 \\
\end{tabular}
\end{subtable}

\begin{subtable}{\linewidth}
\centering
\begin{tabular}{c|cccc}\toprule
\multirow{2}{*}{Activation} & \multicolumn{4}{c}{Products} \\
\cmidrule{2-5}
& 0.005 & 0.01 & 0.05 & 0.1 \\
\midrule
\xmark & 79.2±0.3 & 80.8±0.1 & 83.8±0.2 & 85.0±0.1 \\
\cmark & 84.6±0.0 & 85.8±0.1 & 87.6±0.1 & 88.1±0.0 \\
\bottomrule
\end{tabular}
\end{subtable}

\end{table}

\subsection{Time Efficiency}
\begin{figure}[!t]
\includegraphics[width=\linewidth]{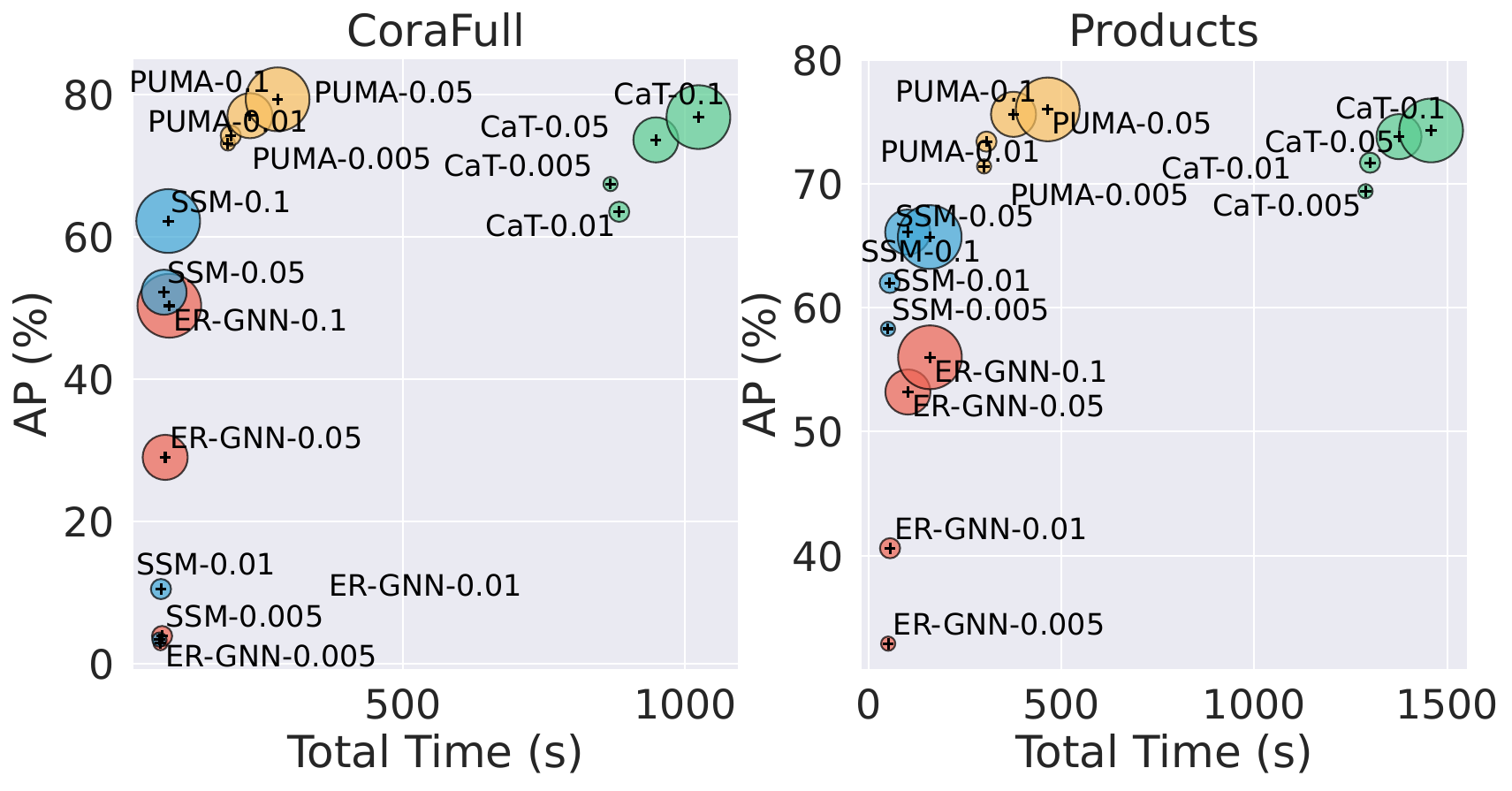}
\caption{Total time (i.e., memorisation time plus training time) for different replay-based CGL methods in the streaming of Arxiv and Products datasets. x-axis represents total time (s) and y-axis indicates the AP (\%). The size of the scatter implies the budget of the memory bank.}
\label{fig:time}
\end{figure}

For replay-based CGL methods, memory generation and model training are two main parts for requiring computation resources. Fig.~\ref{fig:time} compares the total time (memorisation plus training time) for ER-GNN, SSM, CaT and PUMA methods on CoraFull and Products datasets.

For sampling-based methods (e.g., ER-GNN and SSM), memory generation process is efficient. Although the time of graph condensation cannot be ignored as sampling-based methods, the model accuracy is much better than sampling-based methods. For CaT and PUMA, the main computation cost for the memory generation is the feature aggregation in the original graph. Since the condensed graph is significantly smaller than the original graph, the time of feature aggregation in the condensed graph can be ignored.

With the help of an edge-free memory bank, PUMA can ignore the feature aggregation operation during the training phase, allowing the weights of each layer to be learned by an MLP model. As a result, PUMA’s performance can match that of the Joint method, but with faster operation times.



\section{Conclusion}
This paper proposes a novel pseudo-label guided memory bank  (PUMA) method for the CGL problem extending from our CaT~\cite{cat} framework. PUMA enhances the efficacy of condensed graphs by using unlabelled nodes with pseudo labels to leverage more information from the unlabelled nodes. The condensation process efficiency is significantly improved due to one-time propagation and wide graph encoders. Despite balanced sizes in replayed graphs, an imbalance in training losses between existing and new memories is observed, while PUMA addresses this through a retraining strategy. Additionally, PUMA maintains an edge-free memory bank and trains a MLP model for mitigating the costly neighbour message aggregation. In conclusion, PUMA achieves state-of-the-art performance in numerous experiments while maintaining efficiency.



\ifCLASSOPTIONcaptionsoff
  \newpage
\fi



\bibliographystyle{IEEEtranS}
\bibliography{IEEEabrv,ref}

\begin{thebibliography}{10}
\providecommand{\url}[1]{#1}
\csname url@samestyle\endcsname
\providecommand{\newblock}{\relax}
\providecommand{\bibinfo}[2]{#2}
\providecommand{\BIBentrySTDinterwordspacing}{\spaceskip=0pt\relax}
\providecommand{\BIBentryALTinterwordstretchfactor}{4}
\providecommand{\BIBentryALTinterwordspacing}{\spaceskip=\fontdimen2\font plus
\BIBentryALTinterwordstretchfactor\fontdimen3\font minus \fontdimen4\font\relax}
\providecommand{\BIBforeignlanguage}[2]{{%
\expandafter\ifx\csname l@#1\endcsname\relax
\typeout{** WARNING: IEEEtranS.bst: No hyphenation pattern has been}%
\typeout{** loaded for the language `#1'. Using the pattern for}%
\typeout{** the default language instead.}%
\else
\language=\csname l@#1\endcsname
\fi
#2}}
\providecommand{\BIBdecl}{\relax}
\BIBdecl

\bibitem{mas}
R.~Aljundi, F.~Babiloni, M.~Elhoseiny, M.~Rohrbach, and T.~Tuytelaars, ``Memory aware synapses: Learning what (not) to forget,'' in \emph{ECCV}, 2018.

\bibitem{glad}
G.~Cazenavette, T.~Wang, A.~Torralba, A.~A. Efros, and J.~Zhu, ``Generalizing dataset distillation via deep generative prior,'' in \emph{CVPR}, 2023.

\bibitem{herding}
Y.~Chen, M.~Welling, and A.~J. Smola, ``Super-samples from kernel herding,'' in \emph{UAI}, 2010.

\bibitem{dslr}
S.~Choi, W.~Kim, S.~Kim, Y.~In, S.~Kim, and C.~Park, ``{DSLR:} diversity enhancement and structure learning for rehearsal-based graph continual learning,'' in \emph{{WWW}}, 2024.

\bibitem{cgl-kg}
A.~A. Daruna, M.~Gupta, M.~Sridharan, and S.~Chernova, ``Continual learning of knowledge graph embeddings,'' \emph{{IEEE} Robotics Autom. Lett.}, 2021.

\bibitem{mcond}
X.~Gao, T.~Chen, Y.~Zang, W.~Zhang, Q.~V.~H. Nguyen, K.~Zheng, and H.~Yin, ``Graph condensation for inductive node representation learning,'' \emph{ICDE}, 2024.

\bibitem{shortcut}
R.~Geirhos, J.-H. Jacobsen, C.~Michaelis, R.~Zemel, W.~Brendel, M.~Bethge, and F.~A. Wichmann, ``Shortcut learning in deep neural networks,'' \emph{Nature Machine Intelligence}, 2020.

\bibitem{cldc5}
J.~Gu, K.~Wang, W.~Jiang, and Y.~You, ``Summarizing stream data for memory-restricted online continual learning,'' \emph{CoRR}, vol. abs/2305.16645, 2023.

\bibitem{recsys-tkde}
Q.~Guo, F.~Zhuang, C.~Qin, H.~Zhu, X.~Xie, H.~Xiong, and Q.~He, ``A survey on knowledge graph-based recommender systems,'' \emph{TKDE}, 2022.

\bibitem{sage}
W.~L. Hamilton, Z.~Ying, and J.~Leskovec, ``Inductive representation learning on large graphs,'' in \emph{NeurIPS}, 2017.

\bibitem{altopt}
H.~Han, X.~Liu, H.~Mao, M.~Torkamani, F.~Shi, V.~Lee, and J.~Tang, ``Alternately optimized graph neural networks,'' in \emph{ICML}, 2023.

\bibitem{erkd}
T.~D. Hoang, D.~V. Tung, D.-H. Nguyen, B.-S. Nguyen, H.~H. Nguyen, and H.~Le, ``Universal graph continual learning,'' \emph{arXiv preprint arXiv:2308.13982}, 2023.

\bibitem{ogb}
W.~Hu, M.~Fey, M.~Zitnik, Y.~Dong, H.~Ren, B.~Liu, M.~Catasta, and J.~Leskovec, ``Open graph benchmark: Datasets for machine learning on graphs,'' in \emph{NeurIPS}, 2020.

\bibitem{gc-one}
W.~Jin, X.~Tang, H.~Jiang, Z.~Li, D.~Zhang, J.~Tang, and B.~Yin, ``Condensing graphs via one-step gradient matching,'' in \emph{KDD}, 2022.

\bibitem{gc-multi}
W.~Jin, L.~Zhao, S.~Zhang, Y.~Liu, J.~Tang, and N.~Shah, ``Graph condensation for graph neural networks,'' in \emph{ICLR}, 2022.

\bibitem{gcn}
T.~N. Kipf and M.~Welling, ``Semi-supervised classification with graph convolutional networks,'' in \emph{ICLR}, 2017.

\bibitem{ewc}
J.~Kirkpatrick, R.~Pascanu, N.~C. Rabinowitz, J.~Veness, G.~Desjardins, A.~A. Rusu, K.~Milan, J.~Quan, T.~Ramalho, A.~Grabska{-}Barwinska, D.~Hassabis, C.~Clopath, D.~Kumaran, and R.~Hadsell, ``Overcoming catastrophic forgetting in neural networks,'' \emph{CoRR}, vol. abs/1612.00796, 2016.

\bibitem{pl}
D.-H. Lee \emph{et~al.}, ``Pseudo-label: The simple and efficient semi-supervised learning method for deep neural networks,'' in \emph{ICML Workshop}, 2013.

\bibitem{pcconv}
B.~Li, E.~Pan, and Z.~Kang, ``Pc-conv: Unifying homophily and heterophily with two-fold filtering,'' in \emph{AAAI}, 2024.

\bibitem{ggnn}
Y.~Li, D.~Tarlow, M.~Brockschmidt, and R.~S. Zemel, ``Gated graph sequence neural networks,'' in \emph{ICLR}, 2016.

\bibitem{lwf}
Z.~Li and D.~Hoiem, ``Learning without forgetting,'' \emph{TPAMI}, 2018.

\bibitem{tkgr}
K.~Liang, L.~Meng, M.~Liu, Y.~Liu, W.~Tu, S.~Wang, S.~Zhou, and X.~Liu, ``Learn from relational correlations and periodic events for temporal knowledge graph reasoning,'' in \emph{SIGIR}, 2023.

\bibitem{twp}
H.~Liu, Y.~Yang, and X.~Wang, ``Overcoming catastrophic forgetting in graph neural networks,'' in \emph{AAAI}, 2021.

\bibitem{dtgc}
M.~Liu, Y.~Liu, K.~Liang, W.~Tu, S.~Wang, S.~Zhou, and X.~Liu, ``Deep temporal graph clustering,'' \emph{ICLR}, 2024.

\bibitem{gc-dist}
M.~Liu, S.~Li, X.~Chen, and L.~Song, ``Graph condensation via receptive field distribution matching,'' \emph{CoRR}, vol. abs/2206.13697, 2022.

\bibitem{cat}
Y.~Liu, R.~Qiu, and Z.~Huang, ``Cat: Balanced continual graph learning with graph condensation,'' in \emph{ICDM}, 2023.

\bibitem{gcondenser}
Y.~Liu, R.~Qiu, and Z.~Huang, ``Gcondenser: Benchmarking graph condensation,'' \emph{arXiv preprint arXiv:2405.14246}, 2024.

\bibitem{gem}
D.~Lopez{-}Paz and M.~Ranzato, ``Gradient episodic memory for continual learning,'' in \emph{NeurIPS}, 2017.

\bibitem{cldc1}
W.~Masarczyk and I.~Tautkute, ``Reducing catastrophic forgetting with learning on synthetic data,'' in \emph{CVPR Workshop}, 2020.

\bibitem{amazon}
J.~J. McAuley, C.~Targett, Q.~Shi, and A.~van~den Hengel, ``Image-based recommendations on styles and substitutes,'' in \emph{{SIGIR}}, 2015.

\bibitem{corafull}
A.~McCallum, K.~Nigam, J.~Rennie, and K.~Seymore, ``Automating the construction of internet portals with machine learning,'' \emph{Inf. Retr.}, 2000.

\bibitem{fgnn}
R.~Qiu, Z.~Huang, J.~Li, and H.~Yin, ``Exploiting cross-session information for session-based recommendation with graph neural networks,'' \emph{TOIS}, 2020.

\bibitem{gag}
R.~Qiu, H.~Yin, Z.~Huang, and T.~Chen, ``{GAG:} global attributed graph neural network for streaming session-based recommendation,'' in \emph{SIGIR}, 2020.

\bibitem{cldc3}
A.~Rosasco, A.~Carta, A.~Cossu, V.~Lomonaco, and D.~Bacciu, ``Distilled replay: Overcoming forgetting through synthetic samples,'' in \emph{CSSL Workshop}, 2021.

\bibitem{cldc4}
M.~Sangermano, A.~Carta, A.~Cossu, and D.~Bacciu, ``Sample condensation in online continual learning,'' in \emph{IJCNN}, 2022.

\bibitem{coreset2}
O.~Sener and S.~Savarese, ``Active learning for convolutional neural networks: {A} core-set approach,'' in \emph{ICLR}, 2018.

\bibitem{ssrm}
J.~Su, D.~Zou, Z.~Zhang, and C.~Wu, ``Towards robust graph incremental learning on evolving graphs,'' in \emph{ICML}, 2023.

\bibitem{casegnn}
Y.~Tang, R.~Qiu, Y.~Liu, X.~Li, and Z.~Huang, ``Casegnn: Graph neural networks for legal case retrieval with text-attributed graphs,'' in \emph{ECIR}, 2024.

\bibitem{metaclgraph}
A.~Unal, A.~Akg{\"u}l, M.~Kandemir, and G.~Unal, ``Meta continual learning on graphs with experience replay,'' \emph{Transactions on Machine Learning Research}, 2023.

\bibitem{gat}
P.~Velickovic, G.~Cucurull, A.~Casanova, A.~Romero, P.~Li{\`{o}}, and Y.~Bengio, ``Graph attention networks,'' in \emph{ICLR}, 2018.

\bibitem{cafe}
K.~Wang, B.~Zhao, X.~Peng, Z.~Zhu, S.~Yang, S.~Wang, G.~Huang, H.~Bilen, X.~Wang, and Y.~You, ``{CAFE:} learning to condense dataset by aligning features,'' in \emph{CVPR}, 2022.

\bibitem{cl-survey1}
L.~Wang, X.~Zhang, H.~Su, and J.~Zhu, ``A comprehensive survey of continual learning: Theory, method and application,'' \emph{CoRR}, vol. abs/2302.00487, 2023.

\bibitem{dc}
T.~Wang, J.~Zhu, A.~Torralba, and A.~A. Efros, ``Dataset distillation,'' \emph{CoRR}, vol. abs/1811.10959, 2018.

\bibitem{h-survey}
X.~Wang, D.~Bo, C.~Shi, S.~Fan, Y.~Ye, and S.~Y. Philip, ``A survey on heterogeneous graph embedding: methods, techniques, applications and sources,'' \emph{TBD}, 2022.

\bibitem{hgan}
X.~Wang, H.~Ji, C.~Shi, B.~Wang, Y.~Ye, P.~Cui, and P.~S. Yu, ``Heterogeneous graph attention network,'' in \emph{WWW}, 2019.

\bibitem{coreset1}
Z.~Wang and J.~Ye, ``Querying discriminative and representative samples for batch mode active learning,'' in \emph{KDD}, 2013.

\bibitem{cldc2}
F.~Wiewel and B.~Yang, ``Condensed composite memory continual learning,'' in \emph{IJCNN}, 2021.

\bibitem{sgc}
F.~Wu, A.~H.~S. Jr., T.~Zhang, C.~Fifty, T.~Yu, and K.~Q. Weinberger, ``Simplifying graph convolutional networks,'' in \emph{ICML}, 2019.

\bibitem{traffic-tkde}
P.~Xie, M.~Ma, T.~Li, S.~Ji, S.~Du, Z.~Yu, and J.~Zhang, ``Spatio-temporal dynamic graph relation learning for urban metro flow prediction,'' \emph{TKDE}, 2023.

\bibitem{gin}
K.~Xu, W.~Hu, J.~Leskovec, and S.~Jegelka, ``How powerful are graph neural networks?'' in \emph{ICLR}, 2019.

\bibitem{cgl-recsys1}
Y.~Xu, Y.~Zhang, W.~Guo, H.~Guo, R.~Tang, and M.~Coates, ``Graphsail: Graph structure aware incremental learning for recommender systems,'' in \emph{CIKM}, 2020.

\bibitem{sgdd}
B.~Yang, K.~Wang, Q.~Sun, C.~Ji, X.~Fu, H.~Tang, Y.~You, and J.~Li, ``Does graph distillation see like vision dataset counterpart?'' \emph{NeurIPS}, 2024.

\bibitem{roland}
J.~You, T.~Du, and J.~Leskovec, ``Roland: graph learning framework for dynamic graphs,'' in \emph{SIGKDD}, 2022.

\bibitem{cgl-survey}
Q.~Yuan, S.~Guan, P.~Ni, T.~Luo, K.~L. Man, P.~W.~H. Wong, and V.~Chang, ``Continual graph learning: {A} survey,'' \emph{CoRR}, vol. abs/2301.12230, 2023.

\bibitem{protein-tkde}
L.~Zhang, Y.~Jiang, and Y.~Yang, ``Gnngo3d: Protein function prediction based on 3d structure and functional hierarchy learning,'' \emph{TKDE}, 2023.

\bibitem{cglb}
X.~Zhang, D.~Song, and D.~Tao, ``Cglb: Benchmark tasks for continual graph learning,'' in \emph{NeurIPS}, 2022.

\bibitem{ssm}
X.~Zhang, D.~Song, and D.~Tao, ``Sparsified subgraph memory for continual graph representation learning,'' in \emph{ICDM}, 2022.

\bibitem{hpn}
X.~Zhang, D.~Song, and D.~Tao, ``Hierarchical prototype networks for continual graph representation learning,'' \emph{TPAMI}, 2023.

\bibitem{ricci}
X.~Zhang, D.~Song, and D.~Tao, ``Ricci curvature-based graph sparsification for continual graph representation learning,'' \emph{TNNLS}, 2023.

\bibitem{dsa}
B.~Zhao and H.~Bilen, ``Dataset condensation with differentiable siamese augmentation,'' in \emph{ICML}, 2021.

\bibitem{dm}
B.~Zhao and H.~Bilen, ``Dataset condensation with distribution matching,'' in \emph{WACV}, 2023.

\bibitem{dcgm}
B.~Zhao, K.~R. Mopuri, and H.~Bilen, ``Dataset condensation with gradient matching,'' in \emph{ICLR}, 2021.

\bibitem{SFGC}
X.~Zheng, M.~Zhang, C.~Chen, Q.~V.~H. Nguyen, X.~Zhu, and S.~Pan, ``Structure-free graph condensation: From large-scale graphs to condensed graph-free data,'' \emph{NeurIPS}, 2023.

\bibitem{classIL-survey}
D.~Zhou, Q.~Wang, Z.~Qi, H.~Ye, D.~Zhan, and Z.~Liu, ``Deep class-incremental learning: {A} survey,'' \emph{CoRR}, vol. abs/2302.03648, 2023.

\bibitem{ergnn}
F.~Zhou and C.~Cao, ``Overcoming catastrophic forgetting in graph neural networks with experience replay,'' in \emph{AAAI}, 2021.

\end{thebibliography}
%

%

\begin{IEEEbiography}[{\includegraphics[width=1in,height=1.25in,clip,keepaspectratio]{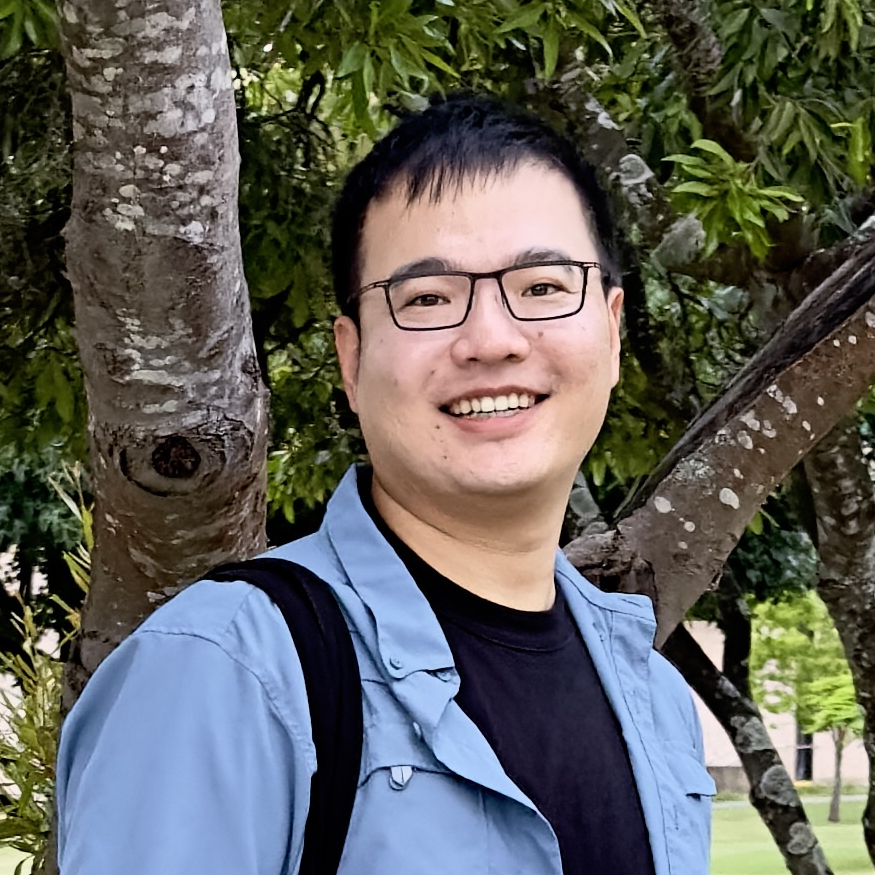}}]{Yilun Liu}
completed his B.E degree at Griffith University in 2020, and M.I.T degree at The University of Queensland in 2022. Currently, he is a PhD student at the School of Electrical Engineering and Computer Science, the University of Queensland. His research interests include graph representation learning, dataset condensation and continual learning.
\end{IEEEbiography}

\begin{IEEEbiography}[{\includegraphics[width=1in,height=1.25in,clip,keepaspectratio]{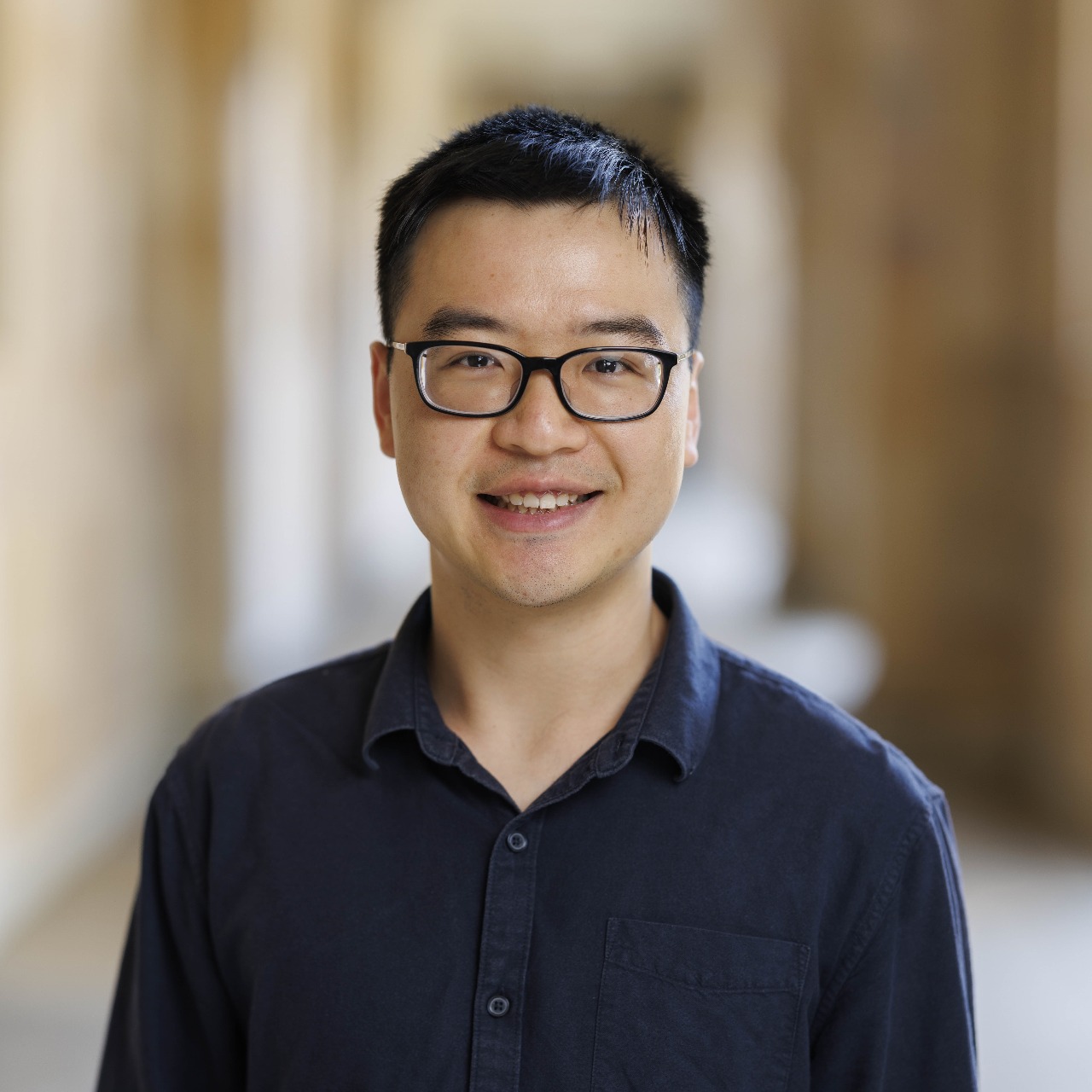}}]{Ruihong Qiu} is currently a Postdoctoral Research Fellow at The University of Queensland. He received his PhD degree in Computer Science at The University of Queensland in 2022. His research mainly focuses on recommender systems, graph neural networks, and data science for cross-disciplinary projects.
\end{IEEEbiography}

\begin{IEEEbiography}[{\includegraphics[width=1in,height=1.25in,clip,keepaspectratio]{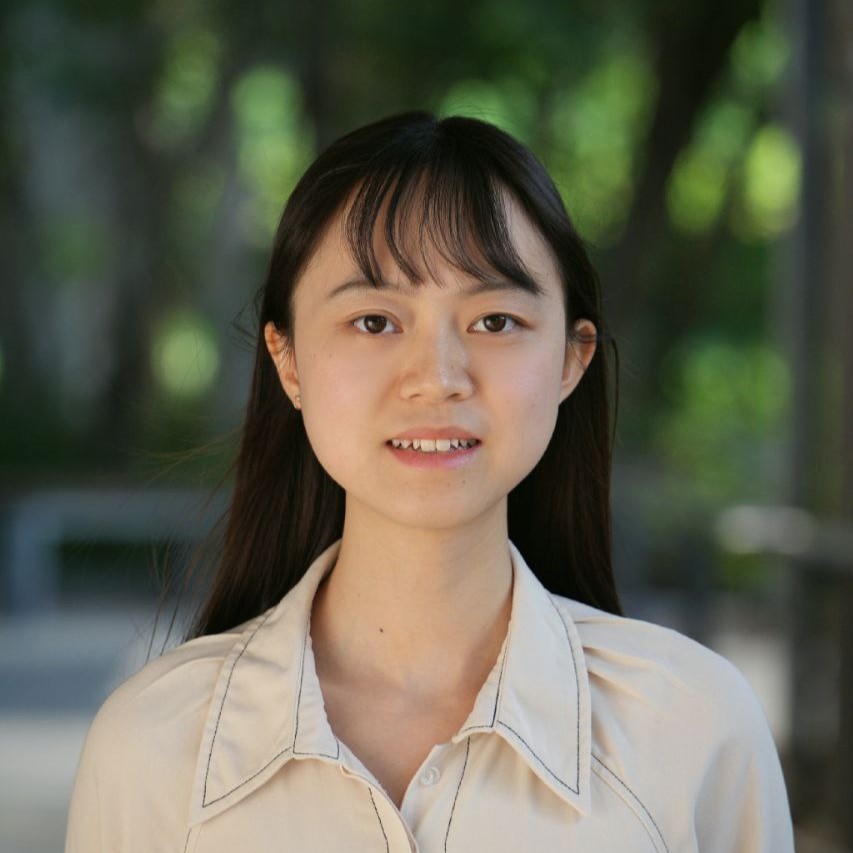}}]{Yanran Tang}
is currently a PhD student at the School of Electrical Engineering and Computer Science, The University of Queensland. She holds an LLB and an LLM degrees. Her research interests include information retrieval and graph representation learning in legal domain.
\end{IEEEbiography}

\begin{IEEEbiography}[{\includegraphics[width=1in,height=1.25in,clip,keepaspectratio]{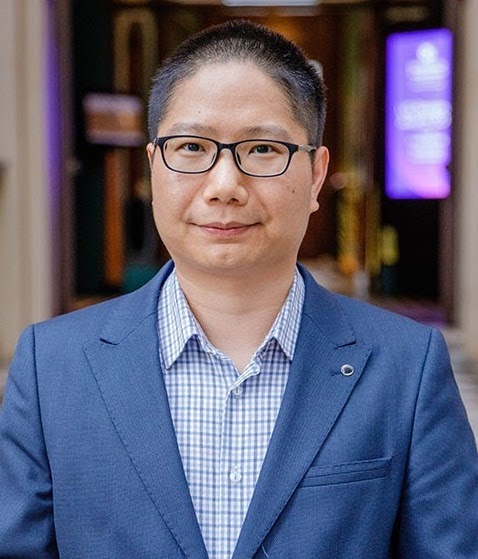}}]{Hongzhi Yin}
received his PhD degree in computer science from Peking University, in 2014. He is a professor and ARC Future Fellow with The University of Queensland. He received the Australian Research Council Discovery Early Career Researcher Award in 2016 and UQ Foundation Research Excellence Award in 2019. His research interests include recommendation system, user profiling, topic models, deep learning, social media mining, and location-based services.
\end{IEEEbiography}

\begin{IEEEbiography}[{\includegraphics[width=1in,height=1.25in,clip,keepaspectratio]{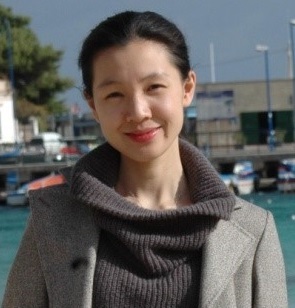}}]{Zi Huang}
received her BSc degree from Tsinghua University in 2001 and her PhD degree in computer science from The University of Queensland (UQ) in 2007. She is a professor and the Discipline Lead for Data Science at the School of Electrical Engineering and Computer Science, UQ. Her research interests mainly include multimedia, computer vision, social data analysis and knowledge discovery. She is currently an associate editor of \textit{The VLDB Journal}, \textit{ACM Transactions on Information Systems} (TOIS), etc and also a member of the VLDB Endowment Board of Trustees.
\end{IEEEbiography}




\end{document}